\documentclass{article}
\usepackage{iclr2027_conference,times}

\usepackage{amsmath,amsfonts,bm}

\def\eqref#1{equation~\ref{#1}}
\def\1{\bm{1}}

\DeclareMathAlphabet{\mathsfit}{\encodingdefault}{\sfdefault}{m}{sl}
\SetMathAlphabet{\mathsfit}{bold}{\encodingdefault}{\sfdefault}{bx}{n}

\usepackage{amsmath,amssymb}
\usepackage{booktabs}
\usepackage{multirow}
\usepackage{graphicx}
\usepackage{wrapfig} % Wrap compact figures and tables with body text.
\usepackage{needspace}
\usepackage{tabularx}
\usepackage{xcolor}
\usepackage{colortbl}
\usepackage{xspace}
\usepackage{enumitem}
\usepackage{microtype}
\usepackage{hyperref}
\usepackage{url}
\usepackage{array}
\usepackage{booktabs}
\usepackage{multirow}
\usepackage{graphicx}
\usepackage{array}
\usepackage{booktabs}
\usepackage{amssymb} % 用于 \vdash 符号
\usepackage{listings}
\usepackage[skins,breakable,most]{tcolorbox}
\usepackage[ruled,vlined]{algorithm2e}

\newcommand{\method}{BoundaryEvo\xspace}

\newcommand{\resultIMOAns}{\textbf{75.3}}
\newcommand{\resultAMO}{\textbf{46.0}}

\newcolumntype{L}[1]{>{\raggedright\arraybackslash}p{#1}}
\newcolumntype{C}[1]{>{\centering\arraybackslash}p{#1}}
\newcolumntype{Y}{>{\centering\arraybackslash}X}

\definecolor{promptframe}{RGB}{40,120,181}
\definecolor{promptback}{RGB}{241,247,251}
\newtcblisting{promptbox}[1]{
  enhanced,
  breakable,
  listing only,
  colframe=promptframe,
  colback=promptback,
  coltitle=white,
  boxrule=0.8pt,
  arc=5pt,
  left=7pt,
  right=7pt,
  top=6pt,
  bottom=6pt,
  title=\textbf{#1},
  fonttitle=\bfseries,
  listing options={
    basicstyle=\ttfamily\footnotesize,
    breaklines=true,
    breakatwhitespace=false,
    columns=fullflexible,
    keepspaces=true,
    showstringspaces=false
  }
}

\title{BoundaryEvo: Data-Model Co-Evolution at the Capability Boundary for Long-Chain \mbox{Reasoning} Distillation}

\author{
Jiaming Su$^{1,*}$, Suorong Yang$^{2,*}$, Xudong Li$^{1,*}$, Tengchao Yang$^{3}$, Peng Ye$^{4,\dagger}$ 
\vspace{2pt}
\\
\small
\begin{tabular}{@{}l@{}}
$^{1}$ Shanghai Artificial Intelligence Laboratory
\quad $^{2}$ Nanjing University
\quad $^{3}$ Tongji University \\[2pt]
$^{4}$ The Chinese University of Hong Kong \\[2pt]
* Equal contribution \quad $^{\dagger}$ Corresponding authors
\end{tabular}
}

\iclrfinalcopy % Uncomment for camera-ready version, but NOT for submission.

\begin{document}
\maketitle

\vspace{-0.2in}
\begin{abstract}
% 这个问题重要
Distilling long-chain reasoning trajectories from large teacher models provides an effective way to improve the reasoning ability of small language models.
% 现有方法的怎么做 忽略了什么
However, existing methods typically select a fixed training set before distillation, overlooking that the usefulness of a reasoning example changes as the student evolves.
% 导致了什么问题
This mismatch becomes especially costly when only a small number of teacher trajectories can be used for training.
Besides, examples already mastered provide little new supervision, while examples far beyond the capability of student may be difficult to learn from.
Training data should therefore adapt as model capability evolves, yet reliable data selection is difficult when the student is initially weak.
% 怎么解决
To address these challenges, our key idea is to first establish the basic reasoning capability and then adapt the training data to the evolving capability boundary of the student.
We propose BoundaryEvo, a two-stage data--model co-evolution framework. 
In \textbf{Capability Priming}, we directly distill a small set of teacher-generated long reasoning trajectories without using a separate model to judge and filter them, allowing the student to acquire an initial reasoning capability before capability-aware data selection.
In \textbf{Capability-Boundary Co-Evolution}, we repeatedly select questions that the current student has not mastered but the expert can reliably solve, and update the training data as the student evolves.
% 实验
With only 1,000 training examples, BoundaryEvo consistently outperforms static and dynamic data selection baselines across 11 reasoning benchmarks, raising the original Qwen3.5-4B accuracy from 62.9 to 75.3 on IMO-AnswerBench and from 26.8 to 46.0 on AMO-Bench.
The code and data will be made publicly available.
% get a repo at https://anonymous.4open.science/
\end{abstract}
\vspace{-0.1in}
\begin{figure}[h]
  \centering
  \includegraphics[width=\linewidth]{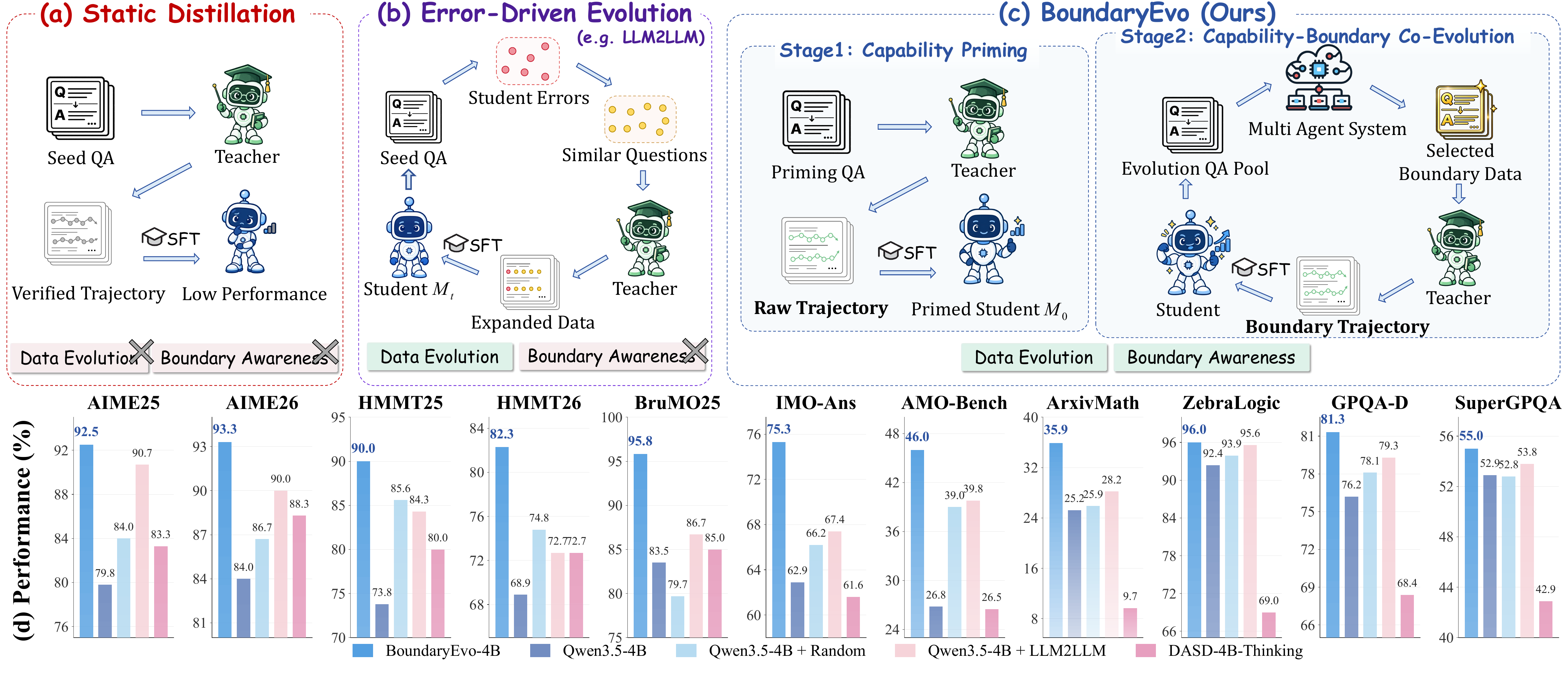}
  \vspace{-0.3in}
  \caption{\textbf{BoundaryEvo evolves supervision along the capability boundary of the student model.} (a-c) Comparison with static distillation and error-driven acquisition, BoundaryEvo dynamically adapts supervision according to the evolving capability boundary of the student model. (d) BoundaryEvo-4B consistently outperforms strong baselines across various reasoning benchmarks.}
  \label{fig:motivation}
\end{figure}

\section{Introduction}
\label{sec:intro}
\vspace{-0.05in}
% 蒸馏很有用 但是在长链遇到了问题太贵了导致数据少 为了解决这个问题
Distilling long-chain reasoning trajectories from large teacher models is an effective way to transfer complex reasoning capabilities to small student models \citep{DBLP:journals/corr/abs-2501-12948,DBLP:journals/corr/abs-2504-16891}.
Yet reasoning supervision is typically constructed \emph{once} \citep{DBLP:conf/acl/HanWFZM025,DBLP:journals/corr/abs-2504-16891} and then treated as static throughout training, even though the learner itself is continuously changing.
As shown in Figure \ref{fig:motivation}(a), existing distillation pipelines typically generate and filter a fixed collection of teacher trajectories before optimization and repeatedly train on the same data thereafter~\citep{DBLP:journals/corr/abs-2601-09088}.
Such a static mechanism overlooks a fundamental property of learning: \emph{data utility is non-stationary with respect to the evolving student}.
Examples that once exposed meaningful capability gaps may later become redundant, whereas trajectories that were initially beyond the student's reach may become precisely the supervision needed for its next stage of improvement.
Thus, the supervision itself should continually adapt to the student's evolving capability boundary, enabling data and model to \textbf{co-evolve} throughout learning.

Recent efforts have begun to adapt training data to the evolving student, as illustrated in Figure~\ref{fig:motivation}(b), typically by prioritizing questions that the current model fails to solve~\citep{DBLP:conf/acl/LeeWKMSAMKG24}.
While this moves beyond static supervision, such adaptation remains largely reactive: it responds to the student's current errors without explicitly modeling how its capability changes over the course of learning. As a result, the training data may shift with the student, but it does not necessarily \emph{co-evolve} with the student's emerging learning needs. For a weak student, failures may span questions that are only slightly beyond its current capability as well as those that remain far out of reach, while student difficulty alone cannot guarantee useful supervision either~\cite{DBLP:journals/tmlr/TzannetosRKS23,DBLP:journals/ml/SongKKL20,chen2025agentfrontierexpandingcapabilityfrontier}.
Thus, data-model co-evolution requires more than tracking what the student gets wrong. It requires continually localizing the \emph{capability boundary} where unresolved student needs align with reliable and informative teacher supervision.

% how we solve, our key idea
Based on this insight, we introduce \textbf{\method}, a two-stage data--model co-evolution framework for long-chain reasoning distillation under a limited supervision budget.
As shown in Figure~\ref{fig:motivation}(c), \method{} organizes reasoning distillation as a two-stage co-evolution process: it first establishes a sufficiently capable student state and subsequently tracks the evolving capability boundary between the student and fixed expert to guide supervision.
In \emph{Stage~1, Capability Priming}, the student is initialized through direct distillation of teacher-generated reasoning trajectories. Rather than specializing supervision to a prematurely estimated learning boundary, this stage exposes the student to a broad distribution of teacher reasoning behaviors, providing the capability foundation required for subsequent boundary-aware adaptation.
In \emph{Stage~2: Capability-Boundary Co-Evolution}, we repeatedly identify questions that remain unresolved by the current student but can be reliably solved by the fixed expert, and then acquire teacher trajectories for the selected questions.
Following each student update, the capability boundary is re-estimated and the supervision set is correspondingly refreshed, forming a closed-loop interaction between model learning and data acquisition.
This iterative coupling progressively redirects supervision toward the student's current learning frontier with reduced supervision budget, allowing the training data and model capability to co-evolve throughout distillation.
With a total budget of only 1,000 training examples, \method consistently outperforms strong static and dynamic data selection baselines on Qwen3.5-4B across 11 reasoning benchmarks (Figure~\ref{fig:motivation}(d)). Compared with the original Qwen3.5-4B model, \method improves accuracy from 62.9 to 75.3 on IMO-AnswerBench and from 26.8 to 46.0 on AMO-Bench. Experiments on Gemma-4-E4B-it further show that the improvements transfer across model backbones.

Our contributions are threefold:
\vspace{-0.1in}
\begin{itemize}[leftmargin=*]
    \item We identify a stage-dependent principle for long-chain reasoning distillation. Early-stage capability priming benefits from retaining diverse teacher reasoning behaviors, including exploration and unsuccessful attempts, rather than filtering trajectories solely by final correctness, while later-stage training benefits from capability-boundary-guided data selection.
    \item We propose \method, a two-stage data--model co-evolution framework for long-chain reasoning distillation, which first establishes student capability through capability priming and then adaptively evolves long-horizon supervision through student--expert capability-boundary tracking. 
    \item We evaluate \method with only 1,000 training examples across 11 reasoning benchmarks and multiple model backbones, showing consistent gains over static and dynamic data-selection baselines and validating the roles of both capability priming and capability-boundary co-evolution.
\end{itemize}

% contributions 不正交 应该是 problem/insight, method, finding

\section{Related Work}
\label{sec:related}
\vspace{-0.05in}

\textbf{Reasoning distillation.}
Knowledge distillation transfers teacher predictions or generated sequences
to student models \citep{DBLP:journals/corr/HintonVD15,DBLP:conf/emnlp/KimR16}.
Chain-of-thought and self-consistency demonstrate the value of intermediate
reasoning trajectories \citep{DBLP:conf/nips/Wei0SBIXCLZ22,DBLP:conf/iclr/0002WSLCNCZ23}.
Recent reasoning models, including DeepSeek-R1 \citep{DBLP:journals/corr/abs-2501-12948}, O1 replication \citep{DBLP:journals/corr/abs-2411-16489}, and DASD \citep{DBLP:journals/corr/abs-2601-09088},
further highlight the effectiveness of long-chain reasoning supervision.
LIMO \citep{DBLP:journals/corr/abs-2502-03387} and s1 \citep{DBLP:conf/emnlp/MuennighoffYSLFHZLCH25} achieve strong performance with small curated datasets, while OpenMathReasoning studies
large-scale mathematical reasoning data construction
\citep{DBLP:journals/corr/abs-2504-16891}.
VibeThinker \citep{DBLP:journals/corr/abs-2606-16140} explores multi-stage reasoning training beyond single-step
distillation.
However, they largely assume a static supervision distribution and overlook the need to adapt training data as model capabilities evolve.

\textbf{Self-evolution and co-evolution.}
Self-improvement methods iteratively generate experience, obtain feedback,
and update models. STaR \citep{DBLP:conf/nips/ZelikmanWMG22} and ReST \citep{DBLP:journals/corr/abs-2308-08998} alternate reasoning generation and training
with feedback signals.
Self-rewarding models improve feedback generation through model-based rewards
\citep{DBLP:conf/icml/YuanPCLSXW24}.
R-Zero \citep{DBLP:journals/corr/abs-2508-05004} , Socratic-Zero \citep{DBLP:journals/corr/abs-2509-24726}, and SeRL \citep{DBLP:conf/nips/FangLZZZCST25} explore self-play and challenger--solver
evolution strategies.
SyncLoop \citep{DBLP:conf/eccv/ChenHLWZCCZZYHXL26} studies multimodal data evolution coupled with model improvement.
Synthetic data approaches, including Self-Instruct \citep{DBLP:conf/acl/WangKMLSKH23}, WizardLM \citep{xu2024evolinstruct}, WizardMath \citep{DBLP:conf/iclr/LuoSX0LTGLCT025},
and MetaMath \citep{DBLP:conf/iclr/YuJSYLZKLWL24}, construct additional instruction or reasoning data.
AutoData \citep{DBLP:journals/corr/abs-2606-25996} introduces an agentic pipeline for task generation and refinement, while LLM2LLM  \citep{DBLP:conf/acl/LeeWKMSAMKG24} selects teacher-generated data
based on student errors.
Different from these methods, BoundaryEvo performs
capability-boundary-driven data--model co-evolution.

\textbf{Data selection.}
Training data selection has been explored through curriculum learning, active learning, and quality-based filtering.
Curriculum learning and active learning connect sample selection with learner
states \citep{DBLP:conf/icml/BengioLCW09,DBLP:journals/csur/RenXCHLGCW22,yang2026dataagentlearningselect}.
LIMA \citep{DBLP:conf/nips/ZhouLX0SMMEYYZG23}, DEITA \citep{DBLP:conf/iclr/0131Z00H24}, QuRating \citep{DBLP:conf/icml/WettigGM024}, and educational-data studies \citep{DBLP:journals/corr/abs-2306-11644} investigate data quality,
diversity, and selection strategies.
D4 \citep{DBLP:conf/nips/TirumalaSAM23} studies deduplication, diversity, and repeated data utilization.
Experience replay provides a general mechanism for retaining previous
knowledge during iterative learning
\citep{DBLP:journals/nature/MnihKSRVBGRFOPB15,DBLP:journals/corr/SchaulQAS15}.
Outcome and process verification introduce additional signals for reasoning
data assessment \citep{DBLP:journals/corr/abs-2110-14168,DBLP:conf/iclr/LightmanKBEBLLS24,
DBLP:journals/corr/abs-2211-14275,DBLP:conf/acl/WangLSXDLCWS24}.
Although model-based judging is useful, it can still be imperfect
\citep{DBLP:conf/nips/ZhengC00WZL0LXZ23}.
BoundaryEvo selects data based on the capability boundary between student and expert, rather than static quality or error signals.

\section{The Proposed BoundaryEvo}
\label{sec:method}
\vspace{-0.05in}

\begin{figure}[t]
  \centering
  \includegraphics[width=\linewidth]{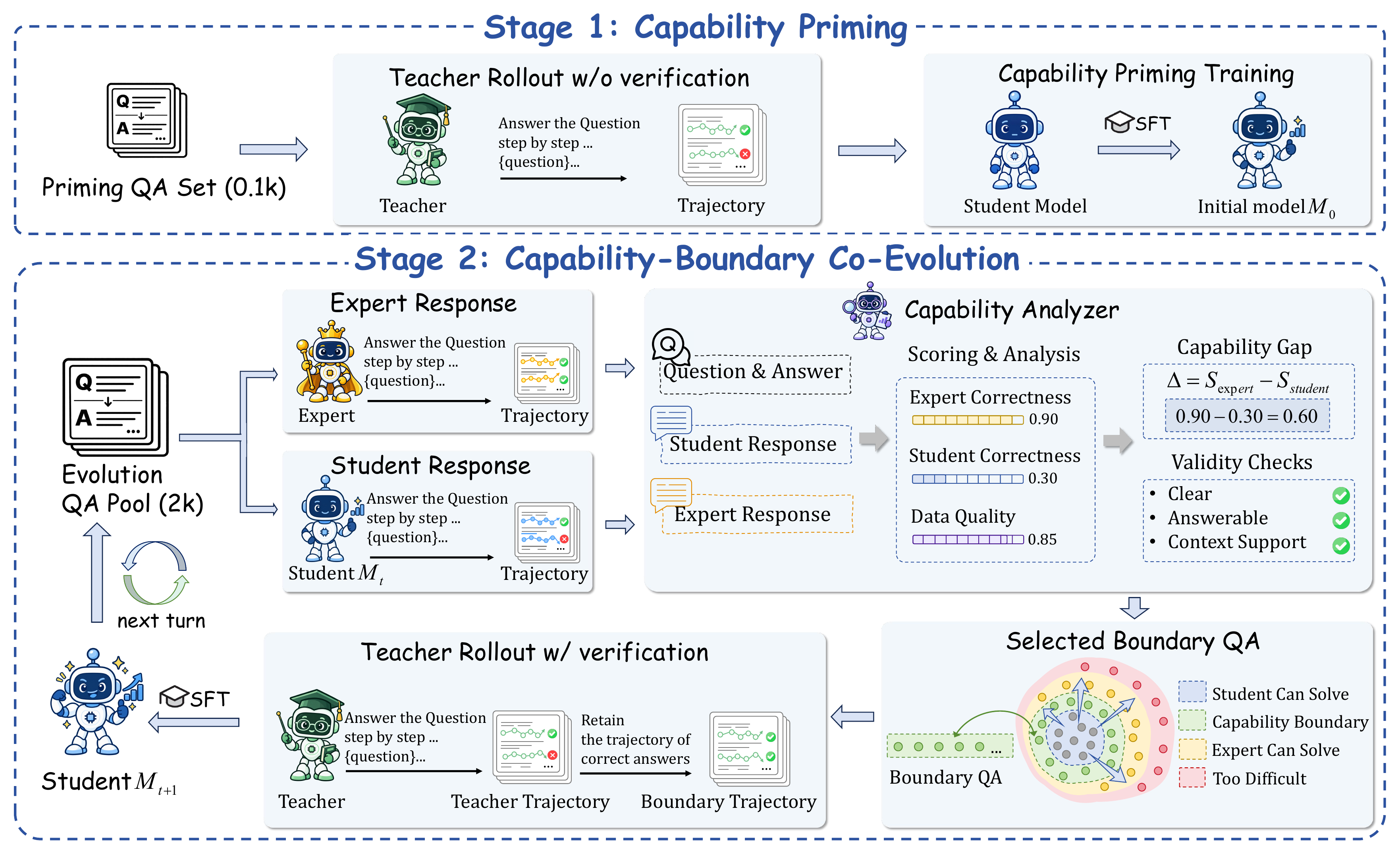}
  \caption{\textbf{Overview of BoundaryEvo.}
  Stage 1 performs judge-free teacher distillation to obtain the capability-primed checkpoint. In Stage 2, responses from the current student and a fixed expert are used to judge the candidate QAs, from which replay and novel records are selected for multiple teacher rollouts. The resulting trajectories are then independently verified against reference answers, and only those that pass verification supervise the next student.}
  \label{fig:pipeline}
  \vspace{-0.1in}
\end{figure}

\textbf{Overview.}
We propose BoundaryEvo, a two-stage data--model co-evolution framework that
dynamically adapts training data to the evolving capabilities of compact
reasoning models. As illustrated in Figure~\ref{fig:pipeline}, Stage~1
performs direct distillation from a small set of teacher-generated reasoning
trajectories without additional model-based judging, providing the student
with foundational reasoning capabilities. Stage~2 then iteratively identifies
capability-boundary examples: at each evolution turn $t$, the current student and a fixed expert model $E$ independently solve the same candidate questions and generate reasoning trajectories, respectively. An analyzer evaluates student performance, expert
solutions, and data quality to select questions that the expert can reliably
solve but the student has not yet mastered. The selected questions are
combined with revalidated historical data, and verified teacher trajectories
are distilled through supervised fine-tuning to update the student model.
As the student improves, the capability boundary shifts and guides subsequent
data selection, enabling continuous data--model co-evolution under a fixed
data budget.

\subsection{Stage 1: Judge-Free Direct Distillation for Capability Priming}
\label{sec:stage1}

Given seed questions $\{x_i\}_{i=1}^{K_0}$, the teacher model generates reasoning trajectories and final answers:
\begin{equation}
    y_i^{\mathrm{T}}\sim T(\cdot\mid x_i), \qquad
    \mathcal{D}_0=\{(x_i,y_i^{\mathrm{T}})\}_{i=1}^{K_0}.
\end{equation}
Stage~1 performs direct reasoning distillation, where teacher-generated
trajectories are used as supervision after basic response-validity checks.
Unlike the subsequent evolution stage, Stage~1 does not apply additional
model-based judging or capability-gap-based filtering. Instead of searching
for difficult examples at this stage, it aims to efficiently transfer teacher reasoning behaviors and establish initial student reasoning capabilities from a small amount of high-quality supervision.

This design avoids premature filtering that may discard useful teacher
experiences due to imperfect automated assessment or diverse valid reasoning
patterns. Training on $\mathcal{D}_0$ produces the capability-primed
checkpoint $M_0$, which provides a stronger initialization for Stage~2.
With this initial capability, BoundaryEvo can more effectively identify
capability-boundary questions and perform subsequent verified teacher
trajectory distillation for iterative data--model co-evolution.

\subsection{Stage 2: Dynamic Capability-Boundary Co-Evolution}
\label{sec:stage2}

\textbf{Student--Expert--Analyzer rollouts and judging.}
At evolution round $t$, the \textbf{student} model $M_{t}$ and a fixed \textbf{expert} independently generate reasoning trajectories for the same candidate question $x_i$:
\begin{equation}
\hat{y}^{\mathrm{s}}_{i,t}\sim M_{t}(\cdot\mid x_i,h_i),
\qquad
\hat{y}^{\mathrm{e}}_i\sim E(\cdot\mid x_i,h_i).
\label{eq:paired-rollouts}
\end{equation}
Here, $h_i$ denotes optional context or auxiliary information. The \textbf{analyzer} evaluates the question, reference information, and reasoning
outputs from both models. The resulting assessment captures three aspects:
\emph{capability-boundary matching}, \emph{data-quality control}, and
\emph{question-validity checks}:
\begin{equation}
\mathbf{a}_i^{(t)}=
\left(
\underbrace{s_i^{\mathrm{s},(t)},s_i^{\mathrm{e}},g_i^{(t)}}_{\text{Boundary matching}},
\underbrace{q_i}_{\text{Quality}},
\underbrace{b_i^{\mathrm{clear}},b_i^{\mathrm{answerable}},c_i}_{\text{Validity}}
\right).
\label{eq:judge-vector}
\end{equation}
The capability gap is computed from the student and expert scores:
\begin{equation}
g_i^{(t)}=s_i^{\mathrm{e}}-s_i^{\mathrm{s},(t)}.
\label{eq:capability-gap}
\end{equation}
A large positive gap indicates that the expert can solve the problem beyond
the current student capability. However, capability difference alone is insufficient, as the expert response must be reliable and the example must provide high-quality and valid supervision. The quality score $q_i$
measures overall training suitability, while validity indicators
$b_i^{\mathrm{clear}}$, $b_i^{\mathrm{answerable}}$, and $c_i$ prevent
ambiguous questions or unsupported contexts from being mistaken as student
limitations.

\textbf{Hard capability-boundary filtering.}
The three levels define the acceptance criteria for capability-boundary
examples. A candidate is selected into $\mathcal{A}_t$ only if it satisfies
all three conditions:
\begin{equation}
\begin{aligned}
\text{Boundary matching:}\quad &
s_i^{\mathrm{e}}\geq\tau_{\mathrm{e}},\quad
s_i^{\mathrm{s},(t)}<\tau_{\mathrm{s}},\quad
g_i^{(t)}\geq\tau_{\mathrm{g}},\\
\text{Quality control:}\quad & q_i\geq\tau_{\mathrm{q}},\\
\text{Question validity:}\quad &
b_i^{\mathrm{clear}}=b_i^{\mathrm{answerable}}=1,\quad
c_i\geq\tau_{\mathrm{c}} .
\end{aligned}
\label{eq:boundary-gate}
\end{equation}
The first condition identifies examples where the expert provides reliable solutions while the student can improve. The remaining conditions ensure that selected samples provide high-quality and valid supervision rather than reflecting ambiguity or data defects. Together, these criteria identify training examples near the current capability boundary. BoundaryEvo therefore selects not only examples that the student fails but also those that can provide reliable learning signals for evolution.

\textbf{Replay and novel acquisition.}
To balance retaining unresolved examples and exploring new boundary samples,
we maintain a fixed training budget of $K=300$ examples per turn. At each
round, the updated student rescans the evolution pool $\mathcal{P}$ of 2K candidates,
producing the accepted set $\mathcal{A}_t$. We divide it into replay examples
previously used in $\mathcal{D}_{t-1}$ and novel examples:
\begin{equation}
    \mathcal R_t=\{z\in\mathcal A_t:x(z)\in
                    \operatorname{Questions}(\mathcal D_{t-1})\},
    \quad
    \mathcal N_t=\{z\in\mathcal A_t:x(z)\notin
                    \operatorname{Questions}(\mathcal D_{t-1})\}.
\label{eq:replay-partition}
\end{equation}
Our replay-first strategy keeps eligible replay examples up to the replay
quota and fills the remaining budget with quality-ranked novel examples:
\begin{equation}
    \mathcal D_t=
    \mathcal R'_t\cup
    \operatorname{Select}_{K-|\mathcal R'_t|}(\mathcal N_t),
    \qquad |\mathcal D_t|=K=300 .
\label{eq:replay-mixture}
\end{equation}
Novel examples are ranked by student response completeness, analyzer
recommendation, and quality score $r_i^{(t)}$, with secondary criteria used
for ties. This design preserves unresolved learning signals while continually
introducing new capability-boundary examples.

\textbf{Post-selection teacher rollouts and verified distillation.}
After QA selection, we construct supervision through fresh teacher rollouts.
This separates \emph{which questions to learn} from
\emph{which solutions to imitate}: the expert rollout identifies the
capability boundary, while the teacher rollouts provide high-quality reasoning trajectories for student training.
For each selected QA record $z_i\in\mathcal D_t$, the teacher independently
samples $L\geq2$ reasoning trajectories:
\begin{equation}
    y_{i,t}^{\mathrm{T},(k)}\sim T(\cdot\mid x_i,h_i),
    \qquad k=1,\ldots,L .
    \label{eq:teacher-multi-rollout}
\end{equation}
We retain trajectories whose final answers pass reference verification:
\begin{equation}
    V(\operatorname{Ans}(y_{i,t}^{\mathrm{T},(k)}),
    y_i^{\mathrm{ref}})=1 .
\end{equation}
A verified trajectory is selected for each question to form
$\mathcal D_t^{\mathrm{distill}}$, and the student is fine-tuned on these
full reasoning trajectories. Unlike Stage~1 judge-free priming, Stage~2
requires evaluation before distillation because data selection now depends
on identifying capability-boundary examples rather than directly transferring
teacher behaviors. QA filtering identifies learning opportunities, while
verification ensures reliable training targets for the next evolution round.
\subsection{Training Objective}

Both stages optimize the student with standard supervised fine-tuning (SFT)
on teacher reasoning trajectories. Let $\mathcal D$ denote
$\mathcal D_0$ in Stage~1 or $\mathcal D_t^{\mathrm{distill}}$ in Stage~2,
where responses include full reasoning and final answers. The objective is
the token-normalized cross-entropy loss:
\begin{equation}
 \mathcal L_{\mathrm{SFT}}(\theta;\mathcal D)=
 -\frac{\sum_{(x_i,y_i)\in\mathcal D}
       \sum_{j=1}^{|y_i|}
       \log p_\theta(y_{ij}\mid x_i,h_i,y_{i,<j})}
       {\sum_{(x_i,y_i)\in\mathcal D}|y_i|}.
\label{eq:sft-objective}
\end{equation}
Only response tokens contribute to the loss. Stage~1 obtains $M_0$ through
direct distillation, while each Stage~2 turn updates $M_{t}$ with verified
capability-boundary trajectories to obtain $M_{t+1}$.

\begin{table*}[t]
\vspace{-0.1in}
\centering
\caption{
Comparison with Reasoning Models under Different Scales and Training Budgets.
Scores are percentages. Data denotes the number of publicly disclosed
reasoning post-training examples and -- indicates unavailable information.
}
\label{tab:main}

\fontsize{7}{8.5}\selectfont
\setlength{\tabcolsep}{1.59pt}
\setlength{\arrayrulewidth}{0.4pt}
\renewcommand{\arraystretch}{1.16}

\begin{tabularx}{\textwidth}{
@{}
L{3.00cm}
C{0.65cm}
C{0.72cm}|
*{8}{Y}|
*{3}{Y}
@{}
}
\toprule
\multirow{3}{*}{\parbox{3.00cm}{\centering\textbf{Model}}}
& \multirow{3}{*}{\parbox{0.65cm}{\centering\textbf{Params}}}
& \multirow{3}{*}{\parbox{0.72cm}{\centering\textbf{Data}}}
& \multicolumn{8}{c|}{\textbf{Mathematical Reasoning}}
& \multicolumn{3}{c}{\textbf{General Reasoning}} \\
\cmidrule(lr){4-11}
\cmidrule(lr){12-14}

& & &
\textbf{AIME} &
\textbf{AIME} &
\textbf{HMMT} &
\textbf{HMMT} &
\textbf{BruMO} &
\textbf{IMO} &
\textbf{AMO} &
\textbf{Arxiv} &
\textbf{Zebra} &
\textbf{GPQA} &
\textbf{Super} \\
[-2pt]
& & &
\textbf{25} & \textbf{26} & \textbf{25} & \textbf{26} & \textbf{25}
& \textbf{-Ans} & \textbf{-Bench} & \textbf{Math} & \textbf{Logic}
& \textbf{-D} & \textbf{GPQA} \\
[-1pt]
\midrule
\rowcolor{gray!15}
\multicolumn{14}{@{}c@{}}{\textit{Small reasoning models}} \\

SmolLM3-3B
& 3B & -- & 36.7 & 41.0 & 26.0 & 36.4 & 49.2 & 28.7
& 5.3 & 7.8 & 26.0 & 41.7 & 29.2 \\

VibeThinker-3B
& 3B & -- & 91.4 & 94.3 & 89.3 & 80.3 & 93.8 & 76.4
& 41.0 & 27.2 & 74.7 & 70.2 & 34.8 \\

DASD-4B-Thinking
& 4B & 448K & 83.3 & 88.3 & 80.0 & 72.7 & 85.0 & 61.6
& 26.5 & 9.7 & 69.0 & 68.4 & 42.9 \\

Hunyuan-4B-Instruct
& 4B & -- & 66.5 & 57.7 & 35.2 & 45.5 & 62.7 & 39.6
& 4.7 & 5.8 & 83.5 & 61.1 & 38.3 \\

Qwen3-4B-Thinking-2507
& 4B & -- & 81.3 & 79.0 & 55.5 & 57.6 & 77.7 & 51.6
& 18.7 & 18.4 & 91.8 & 65.8 & 47.8 \\

Qwen3.5-4B
& 4B & -- & 79.8 & 84.0 & 73.8 & 68.9 & 83.5 & 62.9
& 26.8 & 25.2 & 92.4 & 76.2 & 52.9 \\

NVIDIA-Nemotron-3-Nano
& 4B & -- & 78.5 & -- & -- & -- & -- & --
& -- & -- & -- & 53.2 & -- \\

Spark-X2.5-4B
& 4B & -- & 90.0 & 90.7 & -- & 81.2 & -- & 74.2
& -- & -- & 60.5 & 67.4 & -- \\

Olmo-3-7B-Think
& 7B & $>$102K & 67.9 & 69.1 & 43.8 & 39.4 & 69.0 & 49.4
& 14.0 & 14.6 & 65.0 & 46.2 & -- \\

MiMo-7B-RL-0530
& 7B & 6M & 70.2 & 76.0 & 48.5 & 54.6 & 79.8 & 53.9
& 14.0 & 11.7 & 70.9 & 60.6 & -- \\

OpenReasoning-Nemotron
& 7B & $\geq$5M & 78.2 & 80.2 & 63.5 & 51.5 & 78.8 & 60.6
& 26.0 & 15.5 & 19.8 & 61.1 & -- \\

Gemma-4-12B-it
& 12B & -- & 72.9 & 77.5 & 63.3 & 72.7 & 80.4 & 54.9
& 17.3 & 33.0 & 92.1 & 78.8 & -- \\

Phi-4-reasoning-plus
& 14B & -- & 68.4 & 73.6 & 50.3 & 51.5 & 66.5 & 46.2
& 4.7 & 15.5 & -- & 81.9 & -- \\

Ministral-3-Reasoning-2512
& 14B & -- & 82.9 & 85.0 & 67.1 & 42.4 & 86.7 & 63.4
& 20.7 & 19.4 & 85.1 & 71.2 & -- \\

\midrule
\rowcolor{gray!15}
\multicolumn{14}{@{}c@{}}{\textit{Large reasoning models}} \\

gpt-oss-20b (high)
& 20B & -- & 91.7 & 90.2 & 76.7 & -- & 86.7 & 61.9
& -- & -- & -- & 71.5 & 44.9 \\

NVIDIA-Nemotron-3-Nano
& 30B & -- & 89.1 & 89.9 & 84.6 & -- & -- & 70.4
& -- & -- & -- & 74.1 & -- \\

Nemotron-3.5-Lightning
& 30B & -- & -- & -- & -- & -- & -- & --
& -- & -- & -- & 75.4 & -- \\

GLM-4.7-Flash
& 30B & -- & 91.6 & 88.3 & -- & -- & -- & 73.8
& 53.8 & -- & -- & 75.2 & -- \\

GLM-4.5-Air
& 106B & -- & 83.3 & -- & 69.2 & -- & 90.0 & --
& -- & -- & -- & 75.0 & -- \\

Qwen3-235B-A22B-Thinking
& 235B & -- & 92.3 & -- & 83.9 & -- & -- & 70.5
& 47.8 & -- & 97.5 & 81.1 & 60.7 \\

LongCat-Flash-Thinking
& 560B & -- & 90.6 & -- & 83.7 & -- & -- & --
& 43.6 & -- & 95.5 & 81.5 & -- \\

GPT-5 nano (high)
& N/A & -- & 85.2 & 76.7 & 75.6 & 69.7 & 80.8 & --
& -- & 19.4 & 91.2 & 71.2 & 48.9 \\

\midrule
\rowcolor{gray!15}
\multicolumn{14}{@{}c@{}}{\textit{Our method}} \\

\rowcolor{cyan!5}
\textbf{BoundaryEvo-4B}
& \textbf{4B} & \textbf{1K}
& \textbf{92.5} & \textbf{93.3}
& \textbf{90.0} & \textbf{82.3}
& \textbf{95.8} & \resultIMOAns
& \resultAMO & \textbf{35.9}
& \textbf{96.0} & \textbf{81.3}
& \textbf{55.0} \\

\bottomrule
\end{tabularx}
\vspace{-0.2in}
\end{table*}

\vspace{-0.05in}

\section{Experiments}
\label{sec:experiments}
% \vspace{-0.05in}
\subsection{Experimental Setup}
% \vspace{-0.05in}
\textbf{Models and data.}
We evaluate BoundaryEvo primarily on Qwen3.5-4B
\citep{qwen2026qwen35}, with additional experiments on Qwen3.5-2B \citep{qwen2026qwen35} and
Gemma-4-E4B-it \citep{gemmateam2026gemma4technicalreport}. All QA pairs are sampled from a cleaned subset
(math\_notool.jsonl) of
Nemotron-Cascade-2-SFT-Data \citep{nemotroncascade2sft}, with reasoning trajectories
re-distilled from DeepSeek-V3.2-Speciale \citep{deepseekai2025deepseekv32pushingfrontieropen}. Stage~1 uses 100 teacher
trajectories, while Stage~2 selects from 1,999 candidate questions.
GPT-5.1 \citep{gpt51} provides expert rollouts for capability-boundary estimation, and
Gemini 3.5 Flash \citep{gemini35flash} serves as the analyzer. Across three evolution turns,
BoundaryEvo trains on 1,000 examples in total (100 + 3$\times$300).

\textbf{Training and Evaluation.}
We train models with LoRA-based SFT
\citep{DBLP:conf/iclr/HuSWALWWC22}. Each evolution turn continues from the previous
checkpoint using the selected training trajectories. Training details,
including optimization settings and implementation configurations, are
provided in Appendix~\ref{app:hyperparameters}.
We evaluate reasoning performance on
eight mathematical benchmarks, including AIME25, AIME26 \citep{aime},
HMMT25 \citep{hmmt2025}, HMMT26 \citep{hmmt2026}, BruMO25
\citep{brumo2025}, IMO-AnswerBench (IMO-Ans)
\citep{luong2025robustmathematicalreasoning}, AMO-Bench
\citep{an2025amobenchlargelanguagemodels}, and ArxivMath
\citep{dekoninck2026benchmarksmatharenaevaluationplatform}. We further
measure general reasoning transfer on ZebraLogicBench (ZebraLogic)
\citep{lin2025zebralogicscalinglimitsllms}, GPQA-Diamond (GPQA-D)
\citep{rein2023gpqagraduatelevelgoogleproofqa}, and SuperGPQA
\citep{pteam2025supergpqascalingllmevaluation}. All scores are reported as
percentages, with Math and General denoting unweighted averages over the
corresponding benchmark groups.

\textbf{Baselines.}
We compare zero-shot performance, Stage~1 priming, random selection, and LLM2LLM baseline \citep{DBLP:conf/acl/LeeWKMSAMKG24}. All evolution
methods share the same Stage~1 checkpoint and training budget to isolate the effect of Stage~2 data acquisition. The LLM2LLM baseline selects new
training examples based on student errors under the same training protocol.
We additionally include a static full-data SFT baseline trained on 1,000 examples for comparison with fixed data training.
Besides, Table~\ref{tab:main}
further compares our final model with representative reasoning models across
different parameter scales. The small- and medium-scale group includes
SmolLM3-3B \citep{bakouch2025smollm3}, VibeThinker-3B
\citep{DBLP:journals/corr/abs-2606-16140}, DASD-4B-Thinking
\citep{DBLP:journals/corr/abs-2601-09088}, Hunyuan-4B-Instruct
\citep{hunyuan2025hunyuan4b}, Qwen3-4B-Thinking-2507
\citep{qwen2025qwen34bthinking}, Qwen3.5-4B \citep{qwen2026qwen35},
NVIDIA-Nemotron-3-Nano-4B-BF16 \citep{nvidia2026nemotron3nano4b},
Spark-X2.5-4B \citep{spark2026x25}, Olmo-3-7B-Think
\citep{olmo2026olmo3}, MiMo-7B-RL-0530 \citep{xiaomi2025mimounlockingreasoningpotential},
OpenReasoning-Nemotron-7B \citep{nvidia2025openreasoningnemotron7b},
Gemma-4-12B-it \citep{gemmateam2026gemma4technicalreport}, Phi-4-reasoning-plus
\citep{abdin2025phi4reasoningtechnicalreport}, and Ministral-3-14B-Reasoning-2512
\citep{liu2026ministral3}. The large-scale group includes
gpt-oss-20b \citep{openai2025gptoss120bgptoss20bmodel}, NVIDIA-Nemotron-3-Nano-30B-A3B-BF-16
\citep{nvidia2025nemotron3nanoopen}, NVIDIA-Nemotron-3.5-Lightning-30B-A3B-BF16
\citep{nvidia2026nemotron35lightning}, GLM-4.7-Flash
\citep{5team2025glm45agenticreasoningcoding}, GLM-4.5-Air \citep{5team2025glm45agenticreasoningcoding},
Qwen3-235B-A22B-Thinking \citep{yang2025qwen3technicalreport},
LongCat-Flash-Thinking \citep{meituanlongcatteam2025introducinglongcatflashthinkingtechnicalreport}, and
GPT-5 nano \citep{singh2026openaigpt5card}.

\begin{table*}[t]
\vspace{-0.1in}
\centering
\caption{
Qwen3.5-4B results across three evolution turns. All Stage~2 strategies start from the same Stage~1 checkpoint. Scores are percentages, with Math and General averaged over eight and three benchmarks. Superscripts indicate absolute changes from zero-shot. Bold and underlining mark the highest and second-highest values in each column.
}
\label{tab:main-method-comparison}

\fontsize{7}{8.5}\selectfont
\setlength{\tabcolsep}{0.4pt}
\setlength{\arrayrulewidth}{0.4pt}
\renewcommand{\arraystretch}{1.16}
\definecolor{scoregaincolor}{RGB}{35,133,103}
\definecolor{scorelosscolor}{RGB}{185,70,70}
\providecommand{\scoregain}[2]{#1\makebox[0pt][l]{\hspace{0.05pt}\raisebox{0.55ex}{\normalfont\fontsize{4.5}{4.5}\selectfont\scalebox{0.74}[1]{\textcolor{scoregaincolor}{#2}}}}}
\providecommand{\scoreloss}[2]{#1\makebox[0pt][l]{\hspace{0.05pt}\raisebox{0.55ex}{\normalfont\fontsize{4.5}{4.5}\selectfont\scalebox{0.74}[1]{\textcolor{scorelosscolor}{#2}}}}}
\providecommand{\scoreflat}[2]{#1\makebox[0pt][l]{\hspace{0.05pt}\raisebox{0.55ex}{\normalfont\fontsize{4.5}{4.5}\selectfont\scalebox{0.74}[1]{\textcolor{gray!65}{#2}}}}}

\begin{tabularx}{\textwidth}{
@{}
L{1.3cm}|
*{8}{Y}|
*{3}{Y}|
*{2}{Y}
@{}
}
\toprule
\multirow{3}{*}{\parbox{1.3cm}{\centering\textbf{Method}}}
& \multicolumn{8}{c|}{\textbf{Mathematical Reasoning}}
& \multicolumn{3}{c|}{\textbf{General Reasoning}}
& \multicolumn{2}{c}{\textbf{Avg.}} \\
\cmidrule(lr){2-9}
\cmidrule(lr){10-12}
\cmidrule(lr){13-14}

&
\textbf{AIME} &
\textbf{AIME} &
\textbf{HMMT} &
\textbf{HMMT} &
\textbf{BruMO} &
\textbf{IMO} &
\textbf{AMO} &
\textbf{Arxiv} &
\textbf{Zebra} &
\textbf{GPQA} &
\textbf{Super} &
\multirow{2}{*}{\makebox[0.72cm][c]{\textbf{Math}}} &
\multirow{2}{*}{\makebox[0.72cm][c]{\textbf{General}}}
\\[-2pt]
& \textbf{25} & \textbf{26} & \textbf{25} & \textbf{26} & \textbf{25}
& \textbf{-Ans} & \textbf{-Bench} & \textbf{Math} & \textbf{Logic}
& \textbf{-D} & \textbf{GPQA}
\\[-1pt]
\midrule

\rowcolor{gray!15}
\multicolumn{14}{@{}c@{}}{\textit{Qwen3.5-4B}} \\

+ Zero-shot
& 79.8 & 84.0 & 73.8 & 68.9 & 83.5 & 62.9 & 26.8 & 25.2
& 92.4 & 76.2 & 52.9 & 63.1 & 73.8 \\

+ Stage 1
& \scoregain{84.2}{+4.4} & \scoregain{88.3}{+4.3} & \scoregain{81.7}{+7.9} & \scoregain{69.7}{+0.8}
& \scoregain{85.0}{+1.5} & \scoregain{66.3}{+3.4} & \scoregain{38.5}{+11.7} & \scoregain{28.2}{+3.0}
& \scoregain{92.5}{+0.1} & \scoregain{79.8}{+3.6} & \scoregain{53.4}{+0.5}
& \scoregain{67.7}{+4.6} & \scoregain{75.2}{+1.4} \\

\midrule

\rowcolor{gray!15}
\multicolumn{14}{@{}c@{}}{
  \textit{Stage 2 with Random Selection}
} \\

+ Turn 1
& \scoregain{85.0}{+5.2} & \scoregain{88.3}{+4.3} & \scoregain{82.4}{+8.6} & \scoregain{72.4}{+3.5}
& \scoreloss{80.8}{-2.7} & \scoregain{66.5}{+3.6} & \scoregain{38.5}{+11.7} & \scoregain{27.2}{+2.0}
& \scoregain{92.7}{+0.3} & \scoregain{79.0}{+2.8} & \scoreloss{52.8}{-0.1}
& \scoregain{67.6}{+4.5} & \scoregain{74.8}{+1.0} \\

+ Turn 2
& \scoregain{85.3}{+5.5} & \scoregain{90.6}{+6.6} & \scoregain{82.9}{+9.1} & \scoregain{70.0}{+1.1}
& \scoregain{86.7}{+3.2} & \scoregain{66.8}{+3.9} & \scoregain{38.7}{+11.9} & \scoregain{28.3}{+3.1}
& \scoregain{93.0}{+0.6} & \scoregain{80.2}{+4.0} & \scoreflat{52.9}{+0.0}
& \scoregain{68.7}{+5.6} & \scoregain{75.4}{+1.6} \\

+ Turn 3
& \scoregain{84.0}{+4.2} & \scoregain{86.7}{+2.7} & \scoregain{85.6}{+11.8} & \scoregain{74.8}{+5.9}
& \scoreloss{79.7}{-3.8} & \scoregain{66.2}{+3.3} & \scoregain{39.0}{+12.2} & \scoregain{25.9}{+0.7}
& \scoregain{93.9}{+1.5} & \scoregain{78.1}{+1.9} & \scoreloss{52.8}{-0.1}
& \scoregain{67.7}{+4.6} & \scoregain{74.9}{+1.1} \\

\midrule

\rowcolor{gray!15}
\multicolumn{14}{@{}c@{}}{
  \textit{Stage 2 with LLM2LLM}
} \\

+ Turn 1
& \scoregain{86.0}{+6.2} & \scoregain{89.2}{+5.2} & \scoregain{82.1}{+8.3} & \scoregain{70.8}{+1.9}
& \scoregain{84.8}{+1.3} & \scoregain{66.6}{+3.7} & \scoregain{39.1}{+12.3} & \scoregain{28.4}{+3.2}
& \scoregain{93.3}{+0.9} & \scoregain{77.8}{+1.6} & \scoregain{53.1}{+0.2}
& \scoregain{68.4}{+5.3} & \scoregain{74.7}{+0.9} \\

+ Turn 2
& \scoregain{86.7}{+6.9} & \scoregain{90.7}{+6.7} & \scoregain{83.3}{+9.5} & \scoregain{71.0}{+2.1}
& \scoregain{84.2}{+0.7} & \scoregain{67.0}{+4.1} & \scoregain{38.9}{+12.1} & \scoregain{28.4}{+3.2}
& \scoregain{94.0}{+1.6} & \scoregain{78.8}{+2.6} & \scoregain{53.2}{+0.3}
& \scoregain{68.8}{+5.7} & \scoregain{75.3}{+1.5} \\

+ Turn 3
& \scoregain{\underline{90.7}}{+10.9} & \scoregain{90.0}{+6.0} & \scoregain{84.3}{+10.5} & \scoregain{72.7}{+3.8}
& \scoregain{86.7}{+3.2} & \scoregain{67.4}{+4.5} & \scoregain{39.8}{+13.0} & \scoregain{28.2}{+3.0}
& \scoregain{95.6}{+3.2} & \scoregain{79.3}{+3.1} & \scoregain{53.8}{+0.9}
& \scoregain{70.0}{+6.9} & \scoregain{76.2}{+2.4} \\

\midrule

\rowcolor{gray!15}
\multicolumn{14}{@{}c@{}}{
  \textit{Stage 2 with BoundaryEvo (Ours)}
} \\

+ Turn 1
& \scoregain{89.3}{+9.5} & \scoregain{91.7}{+7.7} & \scoregain{86.0}{+12.2} & \scoregain{75.8}{+6.9}
& \scoregain{85.0}{+1.5} & \scoregain{69.0}{+6.1} & \scoregain{41.5}{+14.7} & \scoregain{\underline{31.1}}{+5.9}
& \scoregain{92.8}{+0.4} & \scoregain{79.3}{+3.1} & \scoregain{\underline{54.3}}{+1.4}
& \scoregain{71.2}{+8.1} & \scoregain{75.5}{+1.7} \\

+ Turn 2
& \scoregain{90.0}{+10.2} & \scoregain{\underline{92.5}}{+8.5} & \scoregain{\underline{86.7}}{+12.9} & \scoregain{\underline{81.8}}{+12.9}
& \scoregain{\underline{90.0}}{+6.5} & \scoregain{\underline{73.3}}{+10.4} & \scoregain{\underline{44.0}}{+17.2} & \scoregain{28.2}{+3.0}
& \scoregain{\underline{95.9}}{+3.5} & \scoregain{\underline{80.8}}{+4.6} & \scoregain{54.1}{+1.2}
& \scoregain{\underline{73.3}}{+10.2} & \scoregain{\underline{76.9}}{+3.1} \\

\rowcolor{cyan!5}
\textbf{+ Turn 3}
& \scoregain{\textbf{92.5}}{+12.7} & \scoregain{\textbf{93.3}}{+9.3} & \scoregain{\textbf{90.0}}{+16.2} & \scoregain{\textbf{82.3}}{+13.4}
& \scoregain{\textbf{95.8}}{+12.3} & \scoregain{\textbf{75.3}}{+12.4} & \scoregain{\textbf{46.0}}{+19.2} & \scoregain{\textbf{35.9}}{+10.7}
& \scoregain{\textbf{96.0}}{+3.6} & \scoregain{\textbf{81.3}}{+5.1} & \scoregain{\textbf{55.0}}{+2.1}
& \scoregain{\textbf{76.4}}{+13.3} & \scoregain{\textbf{77.4}}{+3.6} \\

\bottomrule
\end{tabularx}
\end{table*}

\subsection{Downstream Effectiveness}
\vspace{-0.05in}
BoundaryEvo-4B achieves strong reasoning performance with only 1K training
examples across diverse mathematical benchmarks. As summarized in
Table~\ref{tab:main}, it obtains leading results on AIME25, HMMT25, HMMT26,
BruMO25, and ArxivMath, and consistently outperforms other 4B-scale reasoning
models. Compared with DASD-4B-Thinking, BoundaryEvo improves all shared
benchmarks, including a 19.5-point gain on AMO-Bench, while using only a small
fraction of the training data.

Beyond same-scale comparisons, BoundaryEvo remains competitive with much
larger reasoning models, achieving 75.3 on IMO-AnswerBench and outperforming
Qwen3-235B-A22B-Thinking. It also generalizes to broader reasoning tasks,
reaching 96.0 on ZebraLogicBench, 81.3 on GPQA-Diamond, and 55.0 on
SuperGPQA. These results demonstrate that capability-boundary-guided
data--model co-evolution enables small reasoning models to effectively
extract high-value long-chain reasoning supervision under limited data
budgets.

\subsection{Main Results}
As shown in table~\ref{tab:main-method-comparison}, BoundaryEvo achieves
consistent gains through capability priming and subsequent evolution.
Stage~1 improves the mathematical average from 63.1 to 67.7, and three
evolution turns further increase it to 71.2, 73.3, and 76.4, yielding an
8.7-point gain over priming. The general-reasoning average also improves
from 75.2 to 77.4 after evolution, despite training only on mathematical
data. BoundaryEvo outperforms random selection and LLM2LLM on all eleven
benchmarks, exceeding them by 8.7 and 6.4 points on mathematical averages
and by 2.5 and 1.2 points on general reasoning. The largest gains over
LLM2LLM appear on HMMT26, BruMO25, IMO-AnswerBench, and ArxivMath,
highlighting the advantage of capability-boundary-driven acquisition over
error-based selection. Across evolution turns, random selection quickly
saturates, while LLM2LLM improves more slowly; in contrast, BoundaryEvo
consistently improves across both benchmark groups. These results show that
capability-boundary-driven selection enables progressive improvement through
iterative QA selection and verified trajectory distillation.
% \vspace{-0.1in}

\begin{table*}[t]
\vspace{-0.1in}
\centering
\caption{
Transfer across model size and architecture. Each backbone is primed
before three sequential Stage~2 turns. Scores are percentages.
Superscripts indicate absolute changes from zero-shot. Bold/underline denote the top two values, and blue shading marks the best checkpoint.
}
\label{tab:cross-model-generalization}

\fontsize{7}{8.5}\selectfont
\setlength{\tabcolsep}{0.2pt}
\setlength{\arrayrulewidth}{0.4pt}
\renewcommand{\arraystretch}{1.16}
\definecolor{scoregaincolor}{RGB}{35,133,103}
\definecolor{scorelosscolor}{RGB}{185,70,70}
\providecommand{\scoregain}[2]{#1\makebox[0pt][l]{\hspace{0.3pt}\raisebox{0.65ex}{\normalfont\fontsize{4.2}{4.2}\selectfont\textcolor{scoregaincolor}{#2}}}}
\providecommand{\scoreloss}[2]{#1\makebox[0pt][l]{\hspace{0.3pt}\raisebox{0.65ex}{\normalfont\fontsize{4.2}{4.2}\selectfont\textcolor{scorelosscolor}{#2}}}}

\begin{tabularx}{\textwidth}{
@{}
L{1.3cm}|
*{6}{Y}|
*{2}{Y}|
*{2}{Y}
@{}
}
\toprule
\multirow{3}{*}{\parbox{1.3cm}{\centering\textbf{Method}}}
& \multicolumn{6}{c|}{\textbf{Mathematical Reasoning}}
& \multicolumn{2}{c|}{\textbf{General Reasoning}}
& \multicolumn{2}{c}{\textbf{Avg.}} \\
\cmidrule(lr){2-7}
\cmidrule(lr){8-9}
\cmidrule(lr){10-11}

& \textbf{AIME} &
\textbf{AIME} &
\textbf{HMMT} &
\textbf{HMMT} &
\textbf{BruMO} &
\textbf{IMO} &
\textbf{Zebra} &
\textbf{GPQA} &
\multirow{2}{*}{\makebox[0.72cm][c]{\textbf{Math}}} &
\multirow{2}{*}{\makebox[0.72cm][c]{\textbf{General}}}
\\[-2pt]
& \textbf{25} & \textbf{26} & \textbf{25} & \textbf{26} & \textbf{25}
& \textbf{-Ans} & \textbf{Logic} & \textbf{-D}
\\[-1pt]
\midrule

% ======================== Qwen3.5-2B ========================

\rowcolor{gray!15}
\multicolumn{11}{@{}c@{}}{\textit{Qwen3.5-2B}} \\

 + Zero-shot
& 34.2 & 38.8 & 22.9 & 22.7 & 35.0 & 22.0
& 44.9 & 50.0 & 29.3 & 47.5 \\

 + Stage 1
& \scoregain{38.3}{+4.1} & \scoregain{50.0}{+11.2} & \scoregain{26.7}{+3.8} & \scoregain{24.5}{+1.8}
& \scoregain{43.3}{+8.3} & \scoregain{35.2}{+13.2}
& \scoregain{50.5}{+5.6} & \scoregain{50.5}{+0.5}
& \scoregain{36.3}{+7.0} & \scoregain{50.5}{+3.0} \\

 + Turn 1
& \scoregain{40.0}{+5.8} & \scoregain{\underline{46.7}}{+7.9} & \scoregain{\underline{30.0}}{+7.1} & \scoregain{\textbf{33.3}}{+10.6}
& \scoregain{46.7}{+11.7} & \scoregain{\underline{35.7}}{+13.7}
& \scoregain{63.7}{+18.8} & \scoregain{\underline{52.5}}{+2.5}
& \scoregain{38.7}{+9.4} & \scoregain{58.1}{+10.6} \\

\rowcolor{cyan!5}
\textbf{ + Turn 2}
& \scoregain{\textbf{46.7}}{+12.5} & \scoregain{\textbf{56.7}}{+17.9} & \scoregain{\textbf{33.3}}{+10.4} & \scoregain{27.3}{+4.6}
& \scoregain{\underline{51.7}}{+16.7} & \scoregain{\textbf{35.9}}{+13.9}
& \scoregain{\textbf{68.7}}{+23.8} & \scoregain{\textbf{53.5}}{+3.5}
& \scoregain{\textbf{41.9}}{+12.6} & \scoregain{\textbf{61.1}}{+13.6} \\

 + Turn 3
& \scoregain{\underline{43.3}}{+9.1} & \scoregain{\underline{46.7}}{+7.9} & \scoregain{23.3}{+0.4} & \scoregain{\underline{30.0}}{+7.3}
& \scoregain{\textbf{54.2}}{+19.2} & \scoregain{35.2}{+13.2}
& \scoregain{\underline{63.8}}{+18.9} & \scoregain{51.5}{+1.5}
& \scoregain{\underline{38.8}}{+9.5} & \scoregain{\underline{57.7}}{+10.2} \\

\midrule

% ======================== Gemma-4-E4B-it ========================

\rowcolor{gray!15}
\multicolumn{11}{@{}c@{}}{\textit{Gemma-4-E4B-it}} \\

 + Zero-shot
& 36.7 & 53.3 & 26.7 & 39.4 & 50.8 & 31.6
& 69.4 & 58.6 & 39.8 & 64.0 \\

 + Stage 1
& \scoregain{70.0}{+33.3} & \scoregain{76.7}{+23.4} & \scoregain{50.0}{+23.3} & \scoregain{42.4}{+3.0}
& \scoregain{55.4}{+4.6} & \scoregain{44.9}{+13.3}
& \scoregain{90.1}{+20.7} & \scoregain{60.1}{+1.5}
& \scoregain{56.6}{+16.8} & \scoregain{75.1}{+11.1} \\

 + Turn 1
& \scoregain{\underline{76.7}}{+40.0} & \scoregain{83.3}{+30.0} & \scoregain{\textbf{80.0}}{+53.3} & \scoreloss{33.3}{-6.1}
& \scoregain{55.7}{+4.9} & \scoregain{49.5}{+17.9}
& \scoregain{92.8}{+23.4} & \scoregain{60.6}{+2.0}
& \scoregain{63.1}{+23.3} & \scoregain{76.7}{+12.7} \\

 + Turn 2
& \scoregain{\underline{76.7}}{+40.0} & \scoregain{\textbf{86.7}}{+33.4} & \scoregain{66.7}{+40.0} & \scoregain{\underline{60.6}}{+21.2}
& \scoregain{\underline{60.8}}{+10.0} & \scoregain{\underline{50.3}}{+18.7}
& \scoregain{\underline{94.1}}{+24.7} & \scoregain{\underline{61.1}}{+2.5}
& \scoregain{\underline{67.0}}{+27.2} & \scoregain{\underline{77.6}}{+13.6} \\

\rowcolor{cyan!5}
\textbf{ + Turn 3}
& \scoregain{\textbf{80.0}}{+43.3} & \scoregain{\textbf{86.7}}{+33.4} & \scoregain{\underline{70.0}}{+43.3} & \scoregain{\textbf{66.7}}{+27.3}
& \scoregain{\textbf{67.5}}{+16.7} & \scoregain{\textbf{51.3}}{+19.7}
& \scoregain{\textbf{94.7}}{+25.3} & \scoregain{\textbf{66.2}}{+7.6}
& \scoregain{\textbf{70.4}}{+30.6} & \scoregain{\textbf{80.5}}{+16.5} \\

\bottomrule
\end{tabularx}
\vspace{-0.1in}
\end{table*}

\subsection{Generalization Across Model Backbones}
\vspace{-0.1in}
Table~\ref{tab:cross-model-generalization} evaluates the transferability of
BoundaryEvo across model scales and architectures. For Qwen3.5-2B,
Stage~1 improves the mathematical and general-reasoning averages from 29.3 to
36.3 and from 47.5 to 50.5, respectively. Subsequent boundary-guided evolution improves performance, with Turn~2 achieving the strongest checkpoint
(e.g., AIME25: 46.7, AIME26: 56.7, and ZebraLogicBench: 68.7). However, the
smaller model shows earlier saturation and less stable gains across turns,
suggesting that limited model capacity can constrain long-term evolution. For
Gemma-4-E4B-it, BoundaryEvo achieves more consistent improvements: Stage~1 raises
AIME-25 from 36.7 to 70.0 and ZebraLogicBench from 69.4 to 90.1, while Stage~2
further improves the mathematical and general-reasoning averages to 70.4 and
80.5 at Turn~3. These results demonstrate that BoundaryEvo transfers across
different backbones and adapts its evolution process to varying student
capabilities.
\vspace{-0.1in}

\subsection{Training Dynamics}
\begin{figure*}[t]
  \centering
  \includegraphics[width=\textwidth]{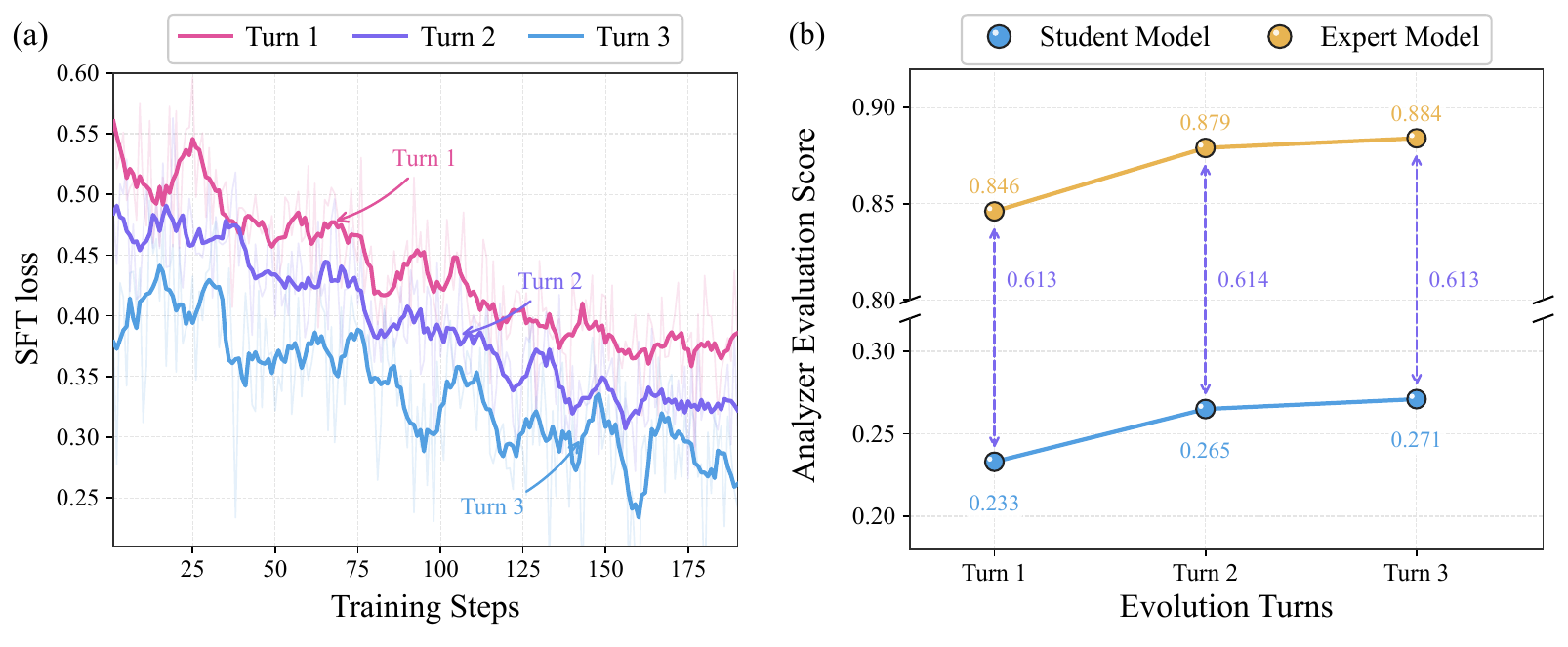}
  \vspace{-0.2in}
  \caption{Training and selection dynamics across the Stage~2.
  (a) SFT loss for representative runs. (b) Mean student and expert scores on each turn. Annotations show the score gap.}
  \label{fig:training-dynamics}
  \vspace{-0.2in}
\end{figure*}

\vspace{-0.05in}
The training dynamics reveal the progressive adaptation enabled by
BoundaryEvo. As illustrated in Figure~\ref{fig:training-dynamics}(a), fitting
the selected teacher trajectories across three evolution turns consistently
reduces the mean loss from 0.4362 to 0.3936 and 0.3388, indicating stable
optimization on newly discovered boundary examples with replay-based
training. As shown in Figure~\ref{fig:training-dynamics}(b), the selected
curriculum gradually shifts as the student improves. The mean student score
increases from 0.233 to 0.265 and 0.271 while expert scores remain high
at 0.846, 0.879, and 0.884, preserving a meaningful student--expert gap
for continued selection of challenging samples. Together, these trends
demonstrate capability-boundary-guided co-evolution, where the curriculum
evolves with student capability while maintaining reliable supervision and
continuously expanding the training frontier.

\subsection{Discussion}
% We conduct ablations to examine the contribution of three key components in
% BoundaryEvo: continued capability-boundary evolution, judge-free capability
% priming, and replay-based training. These studies evaluate whether dynamic
% data evolution, unfiltered initialization, and historical knowledge retention
% are necessary for effective data--model co-evolution.

\begin{wrapfigure}{r}{0.48\textwidth}
% \vspace{-1in}
  \centering
  \includegraphics[width=\linewidth]{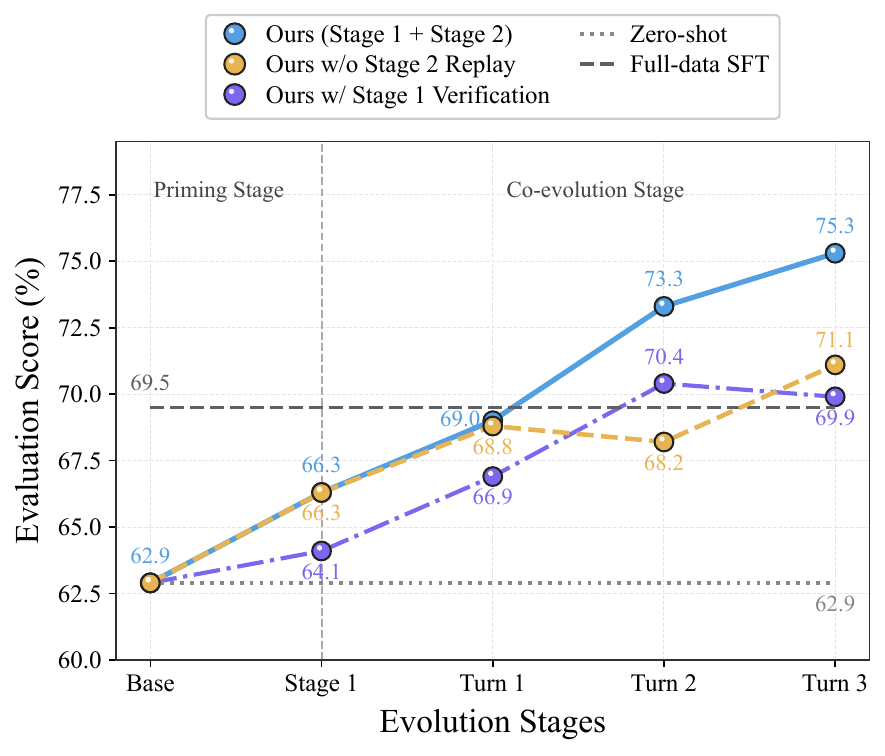}
  \vspace{-0.25in}
  \caption{Ablation study on IMO-AnswerBench performance. The curves compare BoundaryEvo with verified Stage~1 and no-replay variants, while reference lines show zero-shot and static SFT on 1K examples.}
  \label{fig:ablation-trajectory}
  \vspace{-0.2in}
\end{wrapfigure}

\textbf{Does continued evolution improve the training frontier?}
As shown in Figure~\ref{fig:ablation-trajectory}, BoundaryEvo progressively improves IMO-AnswerBench from 66.3 after Stage~1 to 69.0, 73.3, and 75.3 across three evolution turns. The final checkpoint surpasses Stage~1 and static full-data SFT by 9.0 and 5.8 points, respectively. These gains show that continuously updating the training distribution according to the student capability boundary is more effective than using a fixed dataset.

\textbf{Does capability priming benefit from judge-free distillation?}
Table~\ref{tab:ablation_stage1} shows that removing Stage~1 decreases the
final mathematical and general reasoning averages from 76.4/77.4 to
73.5/74.0, demonstrating the importance of capability priming before
evolution. Moreover, verified Stage~1 further reduces performance to
73.1/73.9. As illustrated in Figure~\ref{fig:ablation-trajectory}, direct
distillation starts higher and maintains a larger advantage throughout
evolution. These results suggest that Stage~1 benefits from preserving
diverse teacher reasoning behaviors, while verification is better suited for
Stage~2 boundary refinement.

\begin{wraptable}{r}{0.48\textwidth}
  \vspace{-0.55in}
  \caption{Ablation study on effect of Stage~1 on final Qwen3.5-4B performance. Math and General average eight and three benchmarks.}
  \label{tab:ablation_stage1}
  \fontsize{8}{10.5}\selectfont
  \setlength{\tabcolsep}{4.5pt}
  \begin{tabular}{@{}lcc@{}}
    \toprule
    \textbf{Method} & \textbf{Math AVG} & \textbf{General AVG} \\
    \midrule
    \textit{BoundaryEvo} & 76.4 & 77.4 \\
    \midrule
    % \textit{Ablations} & & \\
    ~~$\vdash$ Stage 1 (w/ verification) & 73.1 & 73.9 \\
    ~~$\vdash$ w/o Stage 1  & 73.5 & 74.0 \\
    \bottomrule
  \end{tabular}
  \vspace{-0.7in}
\end{wraptable}

\textbf{Does replay stabilize co-evolution?} Removing replay reduces IMO-AnswerBench from 75.3 to 71.1 at the final turn. Although newly selected boundary examples improve the student, the performance drop indicates that learning only from the latest curriculum can gradually weaken previously acquired capabilities. Replay therefore helps retain historical knowledge while adapting to newly discovered boundary regions.

Overall, the ablations confirm the complementary roles of BoundaryEvo
components: Stage~1 establishes the initial capability foundation,
boundary-guided evolution expands the training frontier, and replay
stabilizes continual improvement.
\vspace{-0.1in}
\section{Conclusion}
\vspace{-0.1in}
We introduce BoundaryEvo, a dynamic capability-boundary co-evolution
framework for long-chain reasoning distillation. Unlike conventional methods
that treat reasoning data as a static resource, BoundaryEvo continuously
adapts training supervision according to the evolving capabilities of the
student model. It first performs judge-free capability priming to acquire
diverse teacher reasoning behaviors, followed by capability-boundary-guided
data selection and verified teacher trajectory distillation to discover
progressively valuable reasoning supervision. Experiments across 
model backbones demonstrate that BoundaryEvo achieves strong reasoning
performance with only 1K training examples, while ablations validate the
importance of capability priming and capability-boundary-guided evolution. These results show that effective reasoning supervision is not fixed, but can
be continuously discovered through data--model co-evolution.

\clearpage
\subsection*{AI use statement}

We used generative AI tools for limited assistance in language refinement and
writing improvement. We did not use generative AI tools for generating research
ideas, designing experiments, conducting evaluations, or producing scientific
claims. All AI-assisted content was reviewed and revised by the authors, and
the authors take full responsibility for the final content of this work.

\subsection*{Ethics Statement}

This study concerns the training and evaluation of reasoning models. Training
questions are derived from a subset of the publicly released
Nemotron-Cascade-2-SFT-Data. Evaluation uses eleven publicly released
benchmarks: AIME 2025, AIME 2026, HMMT 2025, HMMT 2026, BruMO 2025,
IMO-AnswerBench, AMO-Bench, ArxivMath, ZebraLogicBench, GPQA-Diamond, and
SuperGPQA. These datasets and benchmarks were obtained from their public
sources and used subject to their respective licenses and terms of use. This
work does not recruit human participants or collect private user data, and we
do not intentionally introduce personally identifiable information into the
training or evaluation data.
The empirical comparison reports results for 22 external systems in addition
to BoundaryEvo: SmolLM3-3B, VibeThinker-3B, DASD-4B-Thinking,
Hunyuan-4B-Instruct, Qwen3-4B-Thinking-2507, Qwen3.5-4B,
NVIDIA-Nemotron-3-Nano-4B-BF16, Spark-X2.5-4B, Olmo-3-7B-Think,
MiMo-7B-RL-0530, OpenReasoning-Nemotron-7B, Gemma-4-12B-it,
Phi-4-reasoning-plus, Ministral-3-14B-Reasoning-2512, gpt-oss-20b,
NVIDIA-Nemotron-3-Nano-30B-A3B-BF16,
NVIDIA-Nemotron-3.5-Lightning-30B-A3B-BF16, GLM-4.7-Flash, GLM-4.5-Air,
Qwen3-235B-A22B-Thinking, LongCat-Flash-Thinking, and GPT-5 nano.
Cross-backbone experiments additionally use Qwen3.5-2B and Gemma-4-E4B-it.
Data construction uses DeepSeek-V3.2-Speciale as the trajectory teacher,
GPT-5.1 as the capability-boundary expert, and Gemini 3.5 Flash as the
analyzer. Open checkpoints were accessed from their released sources, while
proprietary models were accessed through authorized APIs under the providers'
terms of service.

\subsection*{Reproducibility statement}

We provide detailed descriptions of the BoundaryEvo framework, data
construction process, training procedure, and evaluation settings in the main
paper and supplementary materials. Implementation details, including
optimization configurations and hyperparameters, are provided in the appendix.
We will release the code and processed data to facilitate reproduction and
future research.



\bibliography{iclr2027_conference}
\bibliographystyle{iclr2027_conference}

\appendix
\clearpage
\onecolumn
\section{Hyperparameter and Selection Sensitivity}
\label{app:sensitivity}

We retain three factors that directly affect the central behavior of
BoundaryEvo: the number of SFT epochs per update, the number of evolution
turns, and the minimum capability gap used by the boundary filter. Other
training settings follow Table~\ref{tab:training-details} and are not repeated
here. We vary one factor at a time from the reference configuration while
holding the backbone, Stage~1 checkpoint, evolution pool, expert, analyzer,
prompts, and per-turn budget fixed.

\begin{table}[!h]
\centering
\caption{Core hyperparameter ablations on Qwen3.5-4B. $E$ is the number of
SFT epochs per update, $R$ is the number of Stage~2 turns, and
$\tau_{\mathrm g}$ is the minimum expert--student capability gap. Stage~2
data count is $300R$ and counts replayed examples whenever they are reused.
Each non-reference row changes one factor only.}
\label{tab:hyperparameter-ablation}
\small
\setlength{\tabcolsep}{7pt}
\renewcommand{\arraystretch}{1.12}
\begin{tabular}{@{}lccccc@{}}
\toprule
\textbf{Ablated factor} & $E$ & $R$ & $\tau_{\mathrm g}$
& \textbf{Stage~2 data} & \textbf{Math / General} \\
\midrule
\rowcolor{cyan!5}
Reference & 5 & 3 & 0.20 & 900 & 76.4 / 77.4 \\
\midrule
Epochs & 1 & 3 & 0.20 & 900 & 72.6 / 75.9 \\
Epochs & 3 & 3 & 0.20 & 900 & 73.5 / 76.1 \\
Epochs & 10 & 3 & 0.20 & 900 & 74.3 / 77.0 \\
\midrule
Evolution turns & 5 & 1 & 0.20 & 300 & 71.2 / 75.5 \\
Evolution turns & 5 & 2 & 0.20 & 600 & 73.3 / 76.9 \\
Evolution turns & 5 & 4 & 0.20 & 1200 & 75.8 / 77.0 \\
\midrule
Minimum gap & 5 & 3 & 0.10 & 900 & 76.1 / 77.2 \\
Minimum gap & 5 & 3 & 0.30 & 900 & 76.2 / 77.2 \\
\bottomrule
\end{tabular}
\end{table}

The epoch sweep tests whether the result depends on under- or over-training
the selected trajectories. The turn sweep measures the benefit of repeatedly
relocating the capability boundary as the student changes. The gap sweep
tests whether selection is robust to admitting easier examples or requiring
a stricter separation between expert and student performance. Together,
these compact controls isolate the most consequential optimization,
evolution, and selection choices.

\section{Training Details}
\label{app:hyperparameters}

This section reports the optimization and implementation details used for
capability priming and the subsequent evolution turns. Unless stated
otherwise, Stage~1 and Stage~2 use the same supervised fine-tuning recipe.
The principal differences across backbones are the conversation template,
the LoRA target modules, and long-sequence preprocessing.

\textbf{Optimization and Parameter-Efficient Fine-Tuning}
We train all student models with response-only LoRA SFT using LLaMA-Factory.
Tokens belonging to the question and system prompt are masked from the loss,
so optimization is applied only to the teacher response. For Qwen3.5, LoRA
is attached to all linear modules. For Gemma-4-E4B-it, it is attached to the
query, key, value, output, gate, up, and down projection modules. Table~\ref{tab:training-details}
summarizes the remaining settings.

\begin{table}[!h]
\centering
\caption{Training configuration used in experiments. The
effective batch size is the product of the per-device batch size, gradient
accumulation steps, and the number of data-parallel workers.}
\label{tab:training-details}
\small
\setlength{\tabcolsep}{7pt}
\renewcommand{\arraystretch}{1.12}
\begin{tabularx}{0.94\textwidth}{@{}L{4.0cm}XX@{}}
\toprule
\textbf{Setting} & \textbf{Qwen3.5-4B / 2B} & \textbf{Gemma-4-E4B-it} \\
\midrule
Training objective & Response-only SFT & Response-only SFT \\
Fine-tuning method & LoRA & LoRA \\
LoRA rank / scaling factor & 64 / 128 & 64 / 128 \\
LoRA target modules & All linear modules & $q,k,v,o$, gate, up, and down projections \\
Trainable parameters & 129.9M / 67.3M & 139.5M \\
Conversation template & \texttt{qwen3} & \texttt{gemma4} \\
Epochs per update & 5 & 5 \\
Optimizer & AdamW, fused PyTorch implementation & Same \\
Learning rate & $1\times10^{-4}$ & $1\times10^{-4}$ \\
Scheduler / warm-up & Cosine / 10\% & Cosine / 10\% \\
Adam $\beta_1,\beta_2,\epsilon$ & $0.9,0.999,10^{-8}$ & Same \\
Weight decay / gradient clipping & $0$ / $1.0$ & Same \\
Numerical precision & bfloat16 & bfloat16 \\
Distributed optimization & DeepSpeed ZeRO-3 & DeepSpeed ZeRO-3 \\
Gradient checkpointing & Enabled & Enabled \\
Per-device batch size & 1 & 1 \\
Default effective batch size & 8 & 8 \\
Maximum trainer sequence length & 40,960 tokens & 40,960 tokens \\
Maximum retained training length & 40,960 tokens & 32,768 tokens \\
Random seed & 42 & 42 \\
\bottomrule
\end{tabularx}
\end{table}

Training jobs use one or four H200 GPUs depending on resource availability while
preserving the default effective batch size of eight. We use gradient
accumulation of 2 for four-GPU jobs and 8 for one-GPU jobs. With 100 Stage~1
examples, five epochs correspond to 65 optimizer updates. A standard
300-example Stage~2 run contains 190 updates.

\textbf{Stage-Wise Initialization and Training Budget}
Stage~1 starts from the original pretrained backbone and trains on 100 direct
teacher trajectories. Its merged checkpoint is denoted by $M_0$. Stage~2 is
sequential: Turn~1 starts from $M_0$, and Turn $t+1$ starts from the merged
student produced by Turn $t$. Each turn contains 300 verified teacher
trajectories selected from the corresponding capability-boundary question
set. Thus, the reported three-turn configuration allocates
$100+3\times300=1000$ training examples. This accounting counts an example
again when it is replayed in a later turn.

After each update, we merge the learned LoRA adapter into its initialization
checkpoint before using the resulting model for evaluation or as the next
turn's student. Consequently, every evolution turn selects data with the
fully updated student rather than with an unmerged adapter or the original
base model. Random-selection and LLM2LLM baselines use the same SFT
objective, optimizer family, epoch count, and backbone-specific formatting.

\textbf{Long-Trajectory Preprocessing}
Training records are stored in OpenAI-style \texttt{messages} format and
contain only fields consumed by the trainer. Qwen3.5 runs use the
\texttt{qwen3} template and a 40,960-token cutoff. For Gemma-4-E4B-it, we render
each record with the native \texttt{gemma4} template and tokenizer before
training. We retain only examples whose rendered length does not exceed
32,768 tokens, while keeping the trainer ceiling at 40,960 tokens. This
separates data selection from implicit truncation and ensures that accepted
Gemma trajectories remain complete. Reasoning markers are converted to the
backbone's native chat representation before tokenization. Across backbones,
malformed records and examples without a usable assistant response are
discarded during preprocessing rather than padded with synthetic content.

\section{Reasoning-Trajectory Length Analysis}
\label{app:reasoning-length}

We test whether the accuracy gains in Table~\ref{tab:main-method-comparison}
require uniformly longer generations. We compare Qwen3.5-4B and the final
BoundaryEvo-4B checkpoint on seven benchmarks, retaining only exact normalized
question matches. Repeated rollouts are summarized by their per-question
median, so every matched item has equal weight. We measure reasoning tokens
with the Qwen tokenizer after removing the final-answer segment and three
thinking-boundary tokens.

\begin{figure}[!ht]
  \centering
  \includegraphics[width=0.98\textwidth]{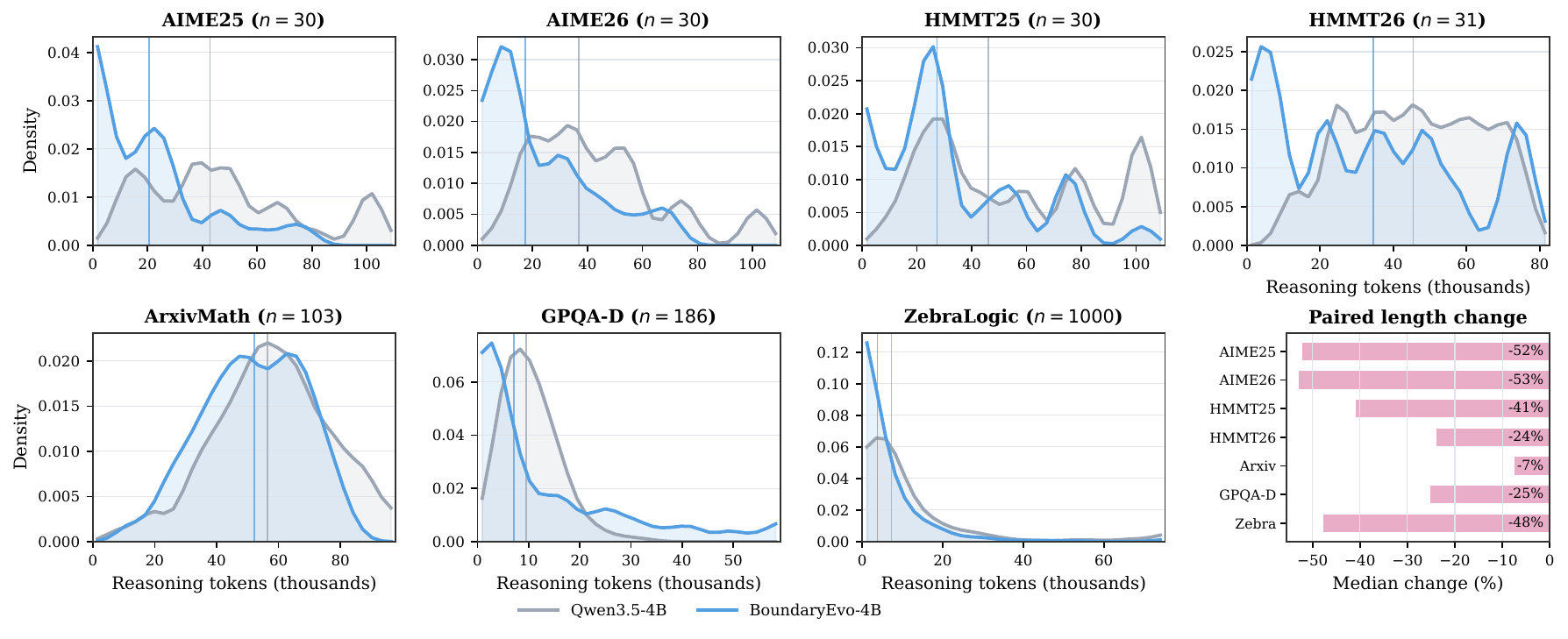}
  \caption{Reasoning-length distributions on matched evaluation questions.
  Curves show smoothed densities of per-question reasoning-token counts, and
  vertical lines mark medians. The final panel reports the relative change in
  median length from Qwen3.5-4B to BoundaryEvo-4B. Repeated rollouts of the
  same question are summarized by their median before plotting.}
  \label{fig:reasoning-length}
\end{figure}

\begin{table}[!ht]
\centering
\caption{Reasoning-length statistics for the matched questions in
Figure~\ref{fig:reasoning-length}. Token counts are reported in thousands.
$\Delta$ denotes the relative change in median length from the base model to
BoundaryEvo-4B.}
\label{tab:reasoning-length}
\small
\setlength{\tabcolsep}{7pt}
\renewcommand{\arraystretch}{1.12}
\begin{tabular}{@{}lrrrrrr@{}}
\toprule
\textbf{Benchmark} & \textbf{$n$} & \multicolumn{2}{c}{\textbf{Median}}
& \textbf{$\Delta$} & \multicolumn{2}{c}{\textbf{90th percentile}} \\
\cmidrule(lr){3-4}\cmidrule(lr){6-7}
& & \textbf{Base} & \textbf{Ours} & & \textbf{Base} & \textbf{Ours} \\
\midrule
AIME25     & 30   & 42.8 & 20.5 & $-52.2\%$ & 99.4  & 52.1 \\
AIME26     & 30   & 36.9 & 17.4 & $-52.9\%$ & 74.9  & 54.9 \\
HMMT25     & 30   & 46.1 & 27.3 & $-40.8\%$ & 102.4 & 74.3 \\
HMMT26     & 31   & 45.4 & 34.6 & $-23.9\%$ & 71.3  & 74.6 \\
ArxivMath  & 103  & 56.4 & 52.2 & $-7.4\%$  & 82.1  & 71.2 \\
GPQA-D     & 186  & 9.5  & 7.1  & $-25.2\%$ & 17.4  & 36.4 \\
ZebraLogic & 1000 & 7.3  & 3.8  & $-47.6\%$ & 25.7  & 16.5 \\
\bottomrule
\end{tabular}
\end{table}

The final model has a lower median reasoning length on every matched
benchmark, with reductions ranging from 7.4\% on ArxivMath to 52.9\% on
AIME26. These reductions coincide with higher accuracy on all seven tasks in
Table~\ref{tab:main-method-comparison}. For example, AIME26 improves from 84.0
to 93.3 while its median trajectory falls from 36.9K to 17.4K tokens, and
ZebraLogic improves from 92.4 to 96.0 while its median falls from 7.3K to
3.8K tokens. The result indicates that the gains do not arise from a uniform
increase in inference length.

Upper-tail behavior remains task dependent. The 90th percentile falls on five
tasks, changes little on HMMT26, and rises on GPQA-D. Thus, BoundaryEvo does
not simply truncate every response. GPQA-D combines a lower median with a
longer tail. Although this observational analysis does not establish
causality, the joint accuracy gains and lower typical lengths are consistent
with more efficient, rather than merely more verbose, reasoning.

% \clearpage
\section{GPQA-D Scaling and Data Efficiency}
\label{app:gpqa-scaling}

We further examine how BoundaryEvo compares with models of different scales
and reported post-training budgets. Both plots in this section use GPQA-D as
the common performance measure and take their values directly from
Table~\ref{tab:main}. We omit systems whose parameter count or GPQA-D result
is unavailable rather than estimating missing values.

\begin{figure}[!ht]
  \centering
  \includegraphics[width=0.92\textwidth]{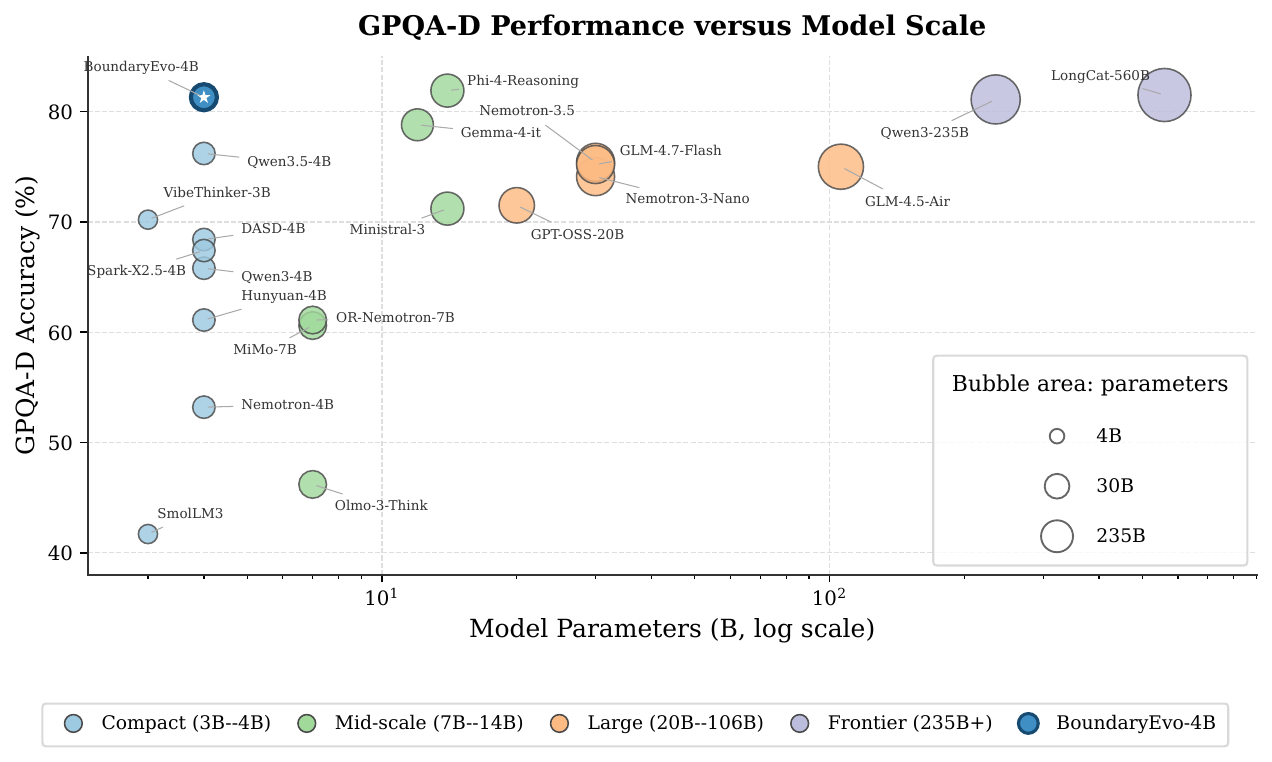}
  \caption{GPQA-D performance versus model scale. The horizontal axis is
  logarithmic, bubble area encodes parameter count, and color denotes the
  parameter range. BoundaryEvo-4B is highlighted with a white star. Only
  models with both a reported parameter count and GPQA-D score in
  Table~\ref{tab:main} are included.}
  \label{fig:model-scale-gpqa}
\end{figure}

Figure~\ref{fig:model-scale-gpqa} shows that strong GPQA-D performance is not
determined by parameter count alone. BoundaryEvo-4B reaches 81.3, improving
over its Qwen3.5-4B base model by 5.1 points and exceeding every reported
model up to 12B parameters. It is also competitive with substantially larger
systems, including Qwen3-235B at 81.1 and LongCat-560B at 81.5, while
Phi-4-Reasoning at 14B obtains the highest displayed score of 81.9. These
results place BoundaryEvo on the upper performance frontier of the plot,
rather than suggesting that it dominates every larger model.

% \clearpage
\begin{figure}[!ht]
  \centering
  \includegraphics[width=0.92\textwidth]{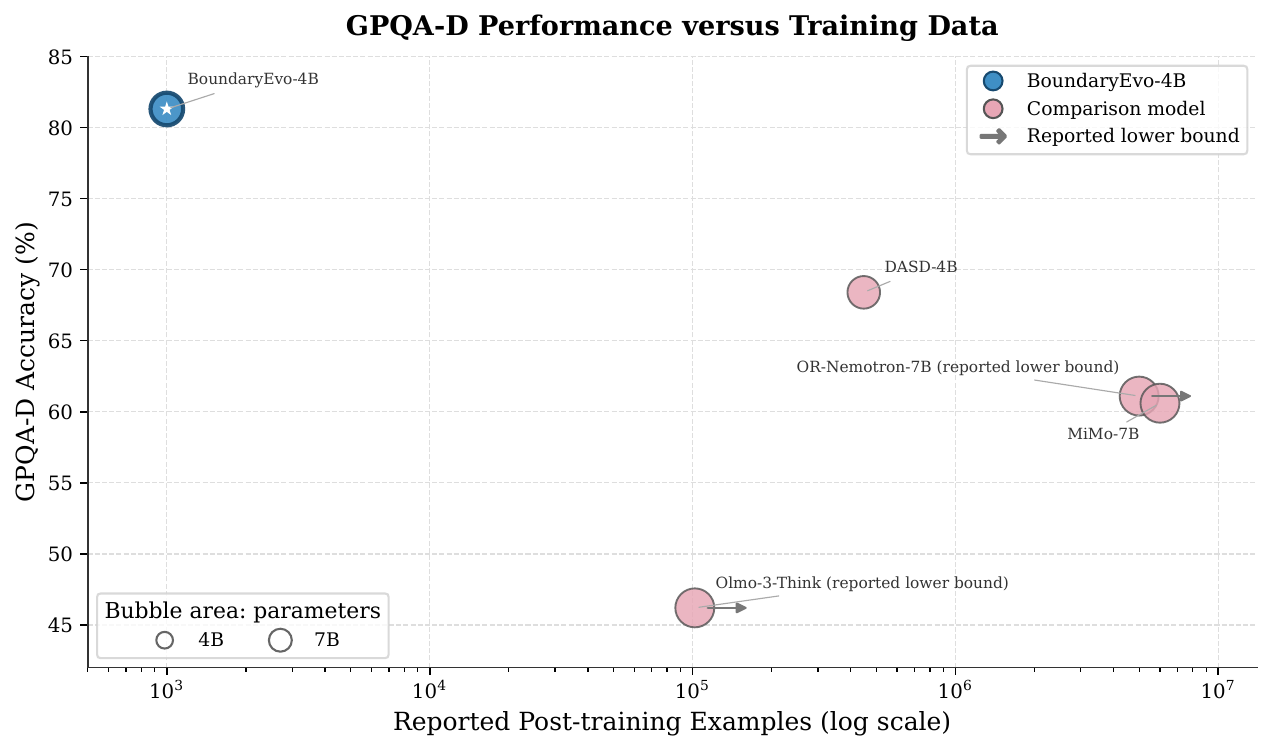}
  \caption{GPQA-D performance versus publicly reported reasoning
  post-training data. The horizontal axis is logarithmic and bubble area
  encodes model parameters. Right-pointing arrows mark reported lower bounds
  rather than exact dataset sizes. Only systems with disclosed data amounts
  and GPQA-D scores in Table~\ref{tab:main} are shown.}
  \label{fig:data-efficiency-gpqa}
\end{figure}

The data-budget view in Figure~\ref{fig:data-efficiency-gpqa} highlights a
larger separation. BoundaryEvo uses 1K cumulative training examples and
scores 81.3. DASD-4B uses 448K reported examples and scores 68.4, while the
three 7B comparison models report at least 102K to 6M examples and score from
46.2 to 61.1. Thus, within this reported set, BoundaryEvo achieves the best
GPQA-D result with the smallest disclosed training set.

This comparison measures example efficiency, not total computational cost.
Dataset definitions, trajectory lengths, and filtering policies differ
across papers. Moreover, BoundaryEvo's 1K count includes repeated training
exposure but excludes student and expert inference, analyzer judging, and
teacher-rollout generation. The plots should therefore be read as a compact
view of reported model and data scale, rather than as a controlled comparison
of training FLOPs.
\section{BoundaryEvo Algorithm and Implementation Details}
\label{app:algorithm}

\textbf{End-to-End Training Procedure.}
Algorithm~\ref{alg:boundaryevo} summarizes the two-stage pipeline, where Stage~1 establishes reasoning capability through judge-free distillation and Stage~2 evolves supervision by selecting capability-boundary questions for verified teacher trajectory distillation.

\begin{algorithm}[H]
\caption{BoundaryEvo: Capability Priming and Capability-Boundary Co-Evolution}
\label{alg:boundaryevo}
\KwIn{Base student $M_{\mathrm{base}}$; trajectory teacher $T$; fixed expert $E$;
analyzer $J$; priming questions $\mathcal Q_0$; evolution pool $\mathcal P$;
reference answers $\{y_i^{\mathrm{ref}}\}$; thresholds $\boldsymbol\tau$;
number of turns $R$; QA budget $K=300$; teacher rollout count $L$}
\KwOut{Evolved student $M_R$}

\tcp{Stage 1: learn how the teacher reasons}
$\mathcal D_0 \leftarrow \emptyset$\;
\ForEach{$x_i \in \mathcal Q_0$}{
  Sample a holistic trajectory $y_i^{\mathrm T}\sim T(\cdot\mid x_i)$\;
  \If{$y_i^{\mathrm T}$ passes basic response-validity checks}{
    $\mathcal D_0 \leftarrow \mathcal D_0\cup\{(x_i,y_i^{\mathrm T})\}$\;
  }
}
$M_0 \leftarrow \operatorname{SFT}(M_{\mathrm{base}},\mathcal D_0)$\;
$\mathcal D_{-1}^{\mathrm{QA}}\leftarrow\emptyset$\;

\tcp{Stage 2: learn what the student should learn next}
\For{$t\leftarrow 0$ \KwTo $R-1$}{
  $\mathcal A_t \leftarrow \emptyset$\;
  \ForEach{$z_i=(x_i,h_i,y_i^{\mathrm{ref}}) \in \mathcal P$}{
    Independently sample $\hat y_{i,t}^{\mathrm s}\sim
    M_t(\cdot\mid x_i,h_i)$ and $\hat y_i^{\mathrm e}\sim
    E(\cdot\mid x_i,h_i)$\;
    $\mathbf a_i^{(t)}\leftarrow J(z_i,\hat y_{i,t}^{\mathrm s},
    \hat y_i^{\mathrm e})$\;
    Compute $g_i^{(t)}=s_i^{\mathrm e}-s_i^{\mathrm s,(t)}$\;
    \If{$s_i^{\mathrm e}\geq\tau_{\mathrm e}$, $s_i^{\mathrm s,(t)}<\tau_{\mathrm s}$,
    $g_i^{(t)}\geq\tau_{\mathrm g}$, $q_i\geq\tau_{\mathrm q}$,
    $b_i^{\mathrm{clear}}=b_i^{\mathrm{answerable}}=1$, and
    $c_i\geq\tau_{\mathrm c}$}{
      $\mathcal A_t \leftarrow \mathcal A_t\cup\{z_i\}$\;
    }
  }
  $\mathcal R_t\leftarrow\{z\in\mathcal A_t:x(z)\in
  \operatorname{Questions}(\mathcal D_{t-1}^{\mathrm{QA}})\}$\;
  $\mathcal N_t\leftarrow\mathcal A_t\setminus\mathcal R_t$\;
  Retain eligible replay examples up to the turn-specific replay quota\;
  Rank novel examples by student-rollout completeness, analyzer recommendation,
  and $r_i^{(t)}$\;
  Fill the remaining slots with ranked novel examples to form
  $\mathcal D_t^{\mathrm{QA}}$, where $|\mathcal D_t^{\mathrm{QA}}|=K$\;

  \tcp{Construct verified supervision only after QA selection}
  $\mathcal D_t^{\mathrm{distill}}\leftarrow\emptyset$\;
  \ForEach{$z_i\in\mathcal D_t^{\mathrm{QA}}$}{
    Sample $\{y_{i,t}^{\mathrm T,(k)}\}_{k=1}^{L}$ independently from
    $T(\cdot\mid x_i,h_i)$\;
    $\mathcal Y_{i,t}^{+}\leftarrow
    \{y:V(\operatorname{Ans}(y),y_i^{\mathrm{ref}})=1\}$\;
    \If{$\mathcal Y_{i,t}^{+}\neq\emptyset$}{
      Select $y_{i,t}^{\star}\in\mathcal Y_{i,t}^{+}$ and add
      $(x_i,h_i,y_{i,t}^{\star})$ to $\mathcal D_t^{\mathrm{distill}}$\;
    }
  }
  $M_{t+1}\leftarrow\operatorname{SFT}(M_t,\mathcal D_t^{\mathrm{distill}})$\;
}
\Return{$M_R$}\;
\end{algorithm}

\textbf{Replay and Training-Set Construction.}
At turn $t$, accepted questions are divided by question identity relative to
the immediately preceding training set. Replay candidates
$\mathcal R_t$ were used in the preceding turn and remain eligible under the
current screening protocol. Novel candidates $\mathcal N_t$ do not occur in
that preceding set. The training budget remains fixed:
\begin{equation}
\mathcal D_t^{\mathrm{QA}}=
\operatorname{Select}_{K_t^{\mathrm R}}(\mathcal R_t)
\cup
\operatorname{Select}_{K_t^{\mathrm N}}(\mathcal N_t),
\qquad K_t^{\mathrm R}+K_t^{\mathrm N}=300.
\end{equation}
Eligible replay questions are retained up to the turn-specific replay quota.

% \clearpage
\section{Prompt Templates}
\label{app:prompts}

We report the raw prompts used by the training-data pipeline. Placeholders in
braces are replaced at runtime. Stage~1 has no model-based judge prompt.

\subsection{Stage 1 Teacher Trajectory Distillation Prompt}

The Stage~1 teacher receives the following system and user messages. The
resulting trajectory is used directly after basic response-validity checks.

\begin{promptbox}{Stage 1 Teacher Trajectory Distillation Prompt}
[System]
You are a helpful and harmless assistant.

You are not allowed to use any tools.

[User]
Please reason step by step, and put your final answer within \boxed{}.

{question}
\end{promptbox}

\subsection{Stage 2 Student and Expert Rollout Prompt}

The current student and fixed expert receive the same prompt and answer the
question independently.

\begin{promptbox}{Stage 2 Student and Expert Rollout Prompt}
Please solve the following problem independently.
Show concise reasoning, then end with exactly one final-answer line:
Final answer: <your answer>

{optional_context}

Problem:
{question}
\end{promptbox}

\subsection{Capability-Boundary Analyzer Prompt}

The analyzer receives the question, optional context, reference answer, and
both independently generated responses. In the raw prompt, \emph{weak} denotes
the current student and \emph{strong} denotes the fixed expert. The implementation also records
reference-correctness and leakage judgments as auxiliary diagnostics. Their
hard rejection switches are disabled in the reported route.

\begin{promptbox}{Capability-Boundary Analyzer Prompt}
[System]
You must output exactly one valid JSON object. Do not include markdown fences, comments, natural-language prefaces, or trailing text.

[User]
You are a strict data-quality and model-trajectory judge.

Evaluate whether this example is suitable training data. Consider:
- whether the question is clear and answerable;
- whether the reference answer is correct;
- if context is provided, whether the answer is supported by the context;
- how good the weak-model answer is;
- how good the strong-model answer is;
- whether the strong answer is meaningfully better than the weak answer.

Scores must be floats in [0, 1].
- weak_score: quality/correctness of the weak answer.
- strong_score: quality/correctness of the strong answer.
- data_quality_score: clarity, answerability, reference correctness,
  and training usefulness.
- context_support_score: use 1.0 when no external context is required
  or no context is provided.

Thresholds used by the pipeline:
{
  "min_strong_score": 0.65,
  "max_weak_score": 0.50,
  "min_gap": 0.20,
  "min_quality_score": 0.70,
  "min_context_support_score": 0.70
}

Question:
{question}

Context:
{context_or_no_context_provided}

Reference answer:
{reference_answer}

Weak model answer:
{student_answer}

Strong model answer:
{expert_answer}

Return exactly one JSON object with this schema:
{
  "is_clear": true,
  "is_answerable": true,
  "is_reference_correct": true,
  "is_context_supported": true,
  "weak_score": 0.25,
  "strong_score": 0.85,
  "data_quality_score": 0.82,
  "context_support_score": 0.9,
  "accepted_recommendation": true,
  "rejection_reasons": [],
  "rationale": "Strong model solves the example correctly, while weak model misses the key reasoning step."
}
\end{promptbox}

\subsection{Final-Answer Verification Prompt}

Each sampled teacher trajectory is judged by final-answer equivalence.
Equivalent algebraic forms and reasonable numerical precision are accepted.

\begin{promptbox}{Final-Answer Verification Prompt}
[System]
You must output exactly one valid JSON object. Do not include markdown fences, comments, natural-language prefaces, or trailing text.

[User]
[JUDGE]
You are a strict math/reasoning answer judge. Decide whether the candidate answer is mathematically equivalent to the reference answer. Allow equivalent forms, algebraic rearrangements, and reasonable numeric precision.

Problem:
{question}

Reference answer:
{reference_answer}

Candidate answer:
{candidate_teacher_answer}

Return exactly one JSON object:
{
  "correct": true,
  "reason": "one concise sentence"
}
\end{promptbox}

\section{Description of Experimental Evaluation}
\label{app:controlled-followup}

\subsection{Common Evaluation Protocol}
\label{app:followup-evaluation}

\textbf{Provenance of reported scores.}
The values in Tables~\ref{tab:main}, \ref{tab:main-method-comparison}, and
\ref{tab:cross-model-generalization} follow a documented source hierarchy.
For model--benchmark pairs reported in the VibeThinker study, official model
cards, or official benchmark leaderboards, we use the corresponding publicly
reported scores. When a model or benchmark combination is not available from
these sources, we perform a local evaluation, primarily for smaller
open-weight models, using the EvalScope framework and the official evaluation
split. These local runs retain EvalScope's default
top-$p$ and temperature and a maximum
of 102,400 generated tokens  setting. The same provenance rule is applied across all three tables.
\textbf{Protocol for controlled follow-up experiments.}
All follow-up methods use the same official evaluation splits, prompts,
answer extractors, decoding settings, evaluator versions, and generation
seed schedule. Competition-math benchmarks use at least four independent completions per question, while all other benchmarks use four. We report
the mean correctness of individual completions rather than majority vote,
best-of-$n$, or pass@$n$. Unparseable and truncated responses without an
extractable final answer are counted as incorrect rather than discarded.
Math and General are the same unweighted benchmark averages used in the
main paper, and the final checkpoint after three turns is fixed as the
primary endpoint.
All methods within a seed
share the priming checkpoint, candidate pool, cached expert outputs, and
available teacher samples. Before training, exact and near-duplicate
questions between the priming/evolution data and every evaluation set are
identified and manually checked; confirmed overlaps are removed from the
common training pool. We will release question-ID manifests, exclusion
counts, and the complete evaluation configuration.

\subsection{Relationship to LLM2LLM and AgentFrontier}
\label{app:followup-baselines}

LLM2LLM~\citep{DBLP:conf/acl/LeeWKMSAMKG24} first identifies student
errors and then asks a teacher to synthesize new examples targeting those
errors. Our current fixed-pool baseline instead rescans the same candidate
pool and selects questions answered incorrectly by the current student.
It isolates error-driven acquisition under the same pool and training
budget, but it is an LLM2LLM-inspired adaptation rather than a complete
reproduction. 
We additionally implement \emph{LLM2LLM-style synthesis}, which retains
the original error-to-new-question step, independently verifies and
de-duplicates the generated QA pairs, and uses the same 300-example budget
per turn. The source revision, prompts, rejected generations, and all
adaptations will be reported.
AgentFrontier~\citep{chen2025agentfrontierexpandingcapabilityfrontier}
also targets tasks that a weaker model cannot solve but a stronger,
tool-augmented system can. BoundaryEvo focuses on a more specific setting:
tool-free long-trajectory distillation for compact students, a fixed QA
pool and expert, and repeated boundary re-estimation after each student
update. Our claimed contribution is therefore the two-stage
priming-and-evolution procedure and its behavior under a small supervision
budget, rather than the general idea of ZPD-style filtering.

% \clearpage
% \onecolumn
% \section{Scaling and Data-Efficiency Analysis}
% \label{app:scaling}

\end{document}